\documentclass[preprint,5p,authoryear]{elsarticle}

\usepackage{amsmath,amssymb,amsfonts}
\usepackage{algpseudocode,algorithm}
\usepackage{subfigure}
\usepackage{graphicx}

\journal{Results in Control and Optimization}

\begin{document}

\sloppy

\begin{frontmatter}



\title{Proximal Policy Optimization with Adaptive Threshold \\for Symmetric Relative Density Ratio}


\author{Taisuke Kobayashi\corref{cor}}
\ead{kobayashi@is.naist.jp}
\ead[url]{http://kbys\_t.gitlab.io/en/}

\cortext[cor]{Corresponding author}

\address{Division of Information Science, Nara Institute of Science and Technology, 8916-5 Takayama-cho, Ikoma, Nara 630-0192, Japan}

\begin{abstract}

Deep reinforcement learning (DRL) is one of the promising approaches for introducing robots into complicated environments.
The recent remarkable progress of DRL stands on regularization of policy, which allows the policy to improve stably and efficiently.
A popular method, so-called proximal policy optimization (PPO), and its variants constrain density ratio of the latest and baseline policies when the density ratio exceeds a given threshold.
This threshold can be designed relatively intuitively, and in fact its recommended value range has been suggested.
However, the density ratio is asymmetric for its center, and the possible error scale from its center, which should be close to the threshold, would depend on how the baseline policy is given.
In order to maximize the values of regularization of policy, this paper proposes a new PPO derived using relative Pearson (RPE) divergence, therefore so-called PPO-RPE, to design the threshold adaptively.
In PPO-RPE, the relative density ratio, which can be formed with symmetry, replaces the raw density ratio.
Thanks to this symmetry, its error scale from center can easily be estimated, hence, the threshold can be adapted for the estimated error scale.
From three simple benchmark simulations, the importance of algorithm-dependent threshold design is revealed.
By simulating additional four locomotion tasks, it is verified that the proposed method statistically contributes to task accomplishment by appropriately restricting the policy updates.

\end{abstract}


\end{frontmatter}


\section{Introduction}

Reinforcement learning (RL)~\citep{sutton2018reinforcement} is one of the promising methodologies for resolving complicated robot control problems.
Remarkable developments have received a lot of attention, especially by grace of the combination with deep neural networks (DNNs)~\citep{lecun2015deep} to approximate value and stochastic policy functions, so-called deep RL (DRL)~\citep{silver2016mastering}.

The basic learning laws for RL have been established with two directions in a relatively early stage~\citep{sutton2018reinforcement}.
One is value-function-based methods, which can find global solutions for the problems with discrete action space.
Another is policy-gradient-based methods, which can handle continuous action space although obtain local solutions.
This paper focuses on the policy-gradient-based methods for applications such as continuous joint control of robots.
Note that these basic learning laws are taken over by DRL.

In the basic research of them, various learning algorithms are being proposed in order to alleviate the numerical instability of the learning process behavior due to nonlinear regression of DNNs:
e.g. experience replay~\citep{lin1992self,schaul2015prioritized};
target network~\citep{mnih2015human,kobayashi2021t};
entropy maximization of policy~\citep{haarnoja2018soft,shi2019soft};
parallelization of learning with asynchronous agents~\citep{mnih2016asynchronous};
and utilization of learned model (i.e. model-based RL)~\citep{chua2018deep,clavera2019model}.
These methods, alone or in combination, have stabilized and accelerated the learning of the given tasks by DRL, and are steadily approaching the stage of practical use.

Among them, research on policy regularization (e.g.~\citep{haarnoja2018soft,shi2019soft}) has recently been systematized theoretically as policy-regularized RL~\citep{geist2019theory}, and its importance can be seen to be high.
The motivation for the policy regularization comes from the fact that the policy improvement is an indirect optimization through the value function, and it is prone to learning instability; in other words, the variance of learning performance due to random seeds is large.
By constraining the update amount and direction of policy, the policy regularization makes the policy smoothly improve and acquire the near-optimal one.

This study further focuses on proximal policy optimization (PPO)~\citep{schulman2017proximal} as one of the policy regularization methods.
The main concept of PPO is proximal update by softly constraining the latest policy to the baseline (basically old) one.
Although PPO has two versions, one if for soft constraint of Kullback-Leibler (KL) divergence and another is for clipping of density ratio, the latter version is major due to its intuitive and practical implementation while outperforming the first version.
In this paper, therefore, PPO refers to the clipping version.

To improve PPO, several methods have been proposed by combining it with other techniques.
For example, \citep{hamalainen2020ppo} optimized the covariance of policy based on CMA-ES to improve the exploration efficiency.
In~\citep{imagawa2019optimistic}, an optimistic bonus according to the uncertainty of return is added to facilitate the exploration.
A single demonstration has been provided in~\citep{libardi2021guided} to accomplish the tasks where rewards are sparsely given.
All of these methods utilize PPO to stabilize learning, and then improve learning efficiency by using other methods in combination.

On the other hand, the way to regularize the policy in PPO is recently revisited.
In PPO-RB~\citep{wang2020truly}, the density ratio over the clipping threshold is explicitly reverted to the clipping range for more certain regularization.
PPOS~\citep{zhu2021functional} relaxes the regularization in PPO-RB by weakening it according to the degree of exceedance from threshold.
In the earlier work of this paper~\citep{kobayashi2021proximal}, a new regularization method has been proposed based on f-divergence to revise the ambiguous regularization caused by clipping.

While PPO has obtained many successes, open issues related to the clipping-based regularization can be raised.
\begin{enumerate}
    \item PPO has no capability to make the latest policy softly constrain to the baseline one~\citep{wang2020truly,zhu2021functional}.
    \item Although the symmetric threshold for clipping is given, the density ratio is in asymmetric domain~\citep{kobayashi2021proximal}.
    \item A recommended threshold range is provided, but it is not for arbitrary learning algorithms.
\end{enumerate}
As for the third problem, a generalized advantage estimation~\citep{schulman2015high} has been utilized for the update law in the original PPO, but the clipping regularization can be acceptable to other learning algorithms.
Although threshold-based regularization achieves the task-invariant tuning, it is easy to imagine that the desirable threshold varies depending on the update speed of policy, which further depends on the learning algorithm combined with the regularization.
The optimization or adaptive design of the threshold for the learning algorithm to be employed has not been proposed yet.
This should be done by considering the error scale from the center of the density ratio (i.e. $1$), but in practice, its asymmetry interferes with the estimation of this error scale.

Hence, this paper proposes a new PPO variant to resolve the above issues by theoretically considering a new regularization problem of relative Pearson (RPE) divergence~\citep{yamada2013relative,sugiyama2013direct}.
Since RPE is one of the divergence metrics between two probabilities, its minimization can softly constrain the latest policy to the baseline one.
To inherit the threshold-based regularization like PPO, the strength of regularization in PPO-RPE is reshaped to the corresponding threshold mathematically.
On the other hand, the relative density ratio is introduced instead of the raw density ratio, and by adjusting its relativity parameter, it can be in symmetric domain.
Thanks to this symmetry, in addition to the earlier work of this paper~\citep{kobayashi2021proximal}, the error scale of the relative density ratio can easily be estimated, and it can be utilized as the adaptive threshold.
Hence, PPO-RPE would be a versatile method that provides mathematically appropriate policy regularization and has a mechanism that automatically adjusts its strength (i.e. threshold) without any consideration of the task and learning algorithm.

As investigation of PPO-RPE, three simple benchmark tasks provided in Pybullet~\citep{coumans2016pybullet} are simulated with two types of learning algorithms.
In their results, the importance of algorithm-dependent threshold design is revealed.
In addition, four locomotion tasks are simulated using PPO and PPO-RPE.
The statistical tests after learning show that only PPO-RPE enables to achieve high scores even in unstable tasks by restricting the policy updates more appropriately than PPO.
Therefore, it can be concluded that PPO-RPE does not require the task- and algorithm-dependent tuning, and brings easy application to a variety of practical problems.

\section{Preliminaries}

\subsection{Policy-regularized reinforcement learning}

Reinforcement learning (RL)~\citep{sutton2018reinforcement} optimizes an agent's policy $\pi$ to maximize the sum of rewards $r$, so-called a return defined at time step $t$ as $R_t = \sum_{k=0}^\infty \gamma^k r_{t+k}$ with $\gamma \in [0, 1)$ discount factor.
This problem can be solved in Markov decision process (MDP) with the tuple $(\mathcal{S}, \mathcal{A}, \mathcal{R}, p_0, p_e)$.
Here, $\mathcal{S}$ and $\mathcal{A}$ denote the state and action spaces, respectively, and $\mathcal{R}$ is the reward set.
$p_0$ and $p_e$ represent the initial state probability and the state transition probability.

Specifically, MDP assumes the following process.
At the time step $t$, the state $s_t \in \mathcal{S}$ is first sampled from environment with either of probabilities, $s_0 \sim p_0(s_0)$ as the initial random state or $s_t \sim p_e(s_t \mid s_{t-1}, a_{t-1})$ as the state transition.
According to $s_t$, an agent decides the action $a_t \in \mathcal{A}$, using the learnable policy $a_t \sim \pi(a_t \mid s_t)$.
$a_t$ acts on the environment and stochastically updates the state to the next one $p_e(s_{t+1} \mid s_t, a_t)$.
At that time, the agent obtains a reward $r_t \in \mathcal{R}$ from the environment: $r_t = r(s_t, a_t)$.
By repeating this process with the experienced data $(s_t, a_t, s_{t+1}, r_t)$, the agent estimates the expected $R_t$ as a value function.
In on-policy learning like PPO~\citep{schulman2017proximal}, the policy-dependent action value function $Q^\pi(s_t, a_t) = \mathbb{E}[R_t \mid s_t, a_t]$ and/or the policy-dependent action value function $V^\pi(s_t) = \mathbb{E}_{a_t \sim \pi}[Q^\pi(s_t, a_t)]$ can be learned through Bellman equation as the following minimization problem:
\begin{align}
    \mathcal{L}(V^\pi) = \frac{1}{2}\left \{ r_t + \gamma V^\pi(s_{t+1}) - V^\pi(s_t) \right \}^2 \to \min
    \label{eq:loss_value}
\end{align}
This paper utilizes $V^\pi$, although $Q^\pi$ can be learned by replacing the above $V^\pi$ with $Q^\pi$.
Note that, with DNNs, the target network is often employed to output $V^\pi(s_{t+1})$ for numerical stability.

The optimization of $\pi$ is formulated as the following minimization problem.
\begin{align}
    \mathcal{L}(\pi) &= - \mathbb{E}_{a_t \sim \pi} \left [ Q^\pi(s_t, a_t) - V^\pi(s_t) \right ]
    \nonumber \\
    &= - \mathbb{E}_{a_t \sim \pi} \left [ A(s_t, a_t) \right ]
    \to \min
    \label{eq:loss_std}
\end{align}
where $A(s_t, a_t) = Q^\pi(s_t, a_t) - V^\pi(s_t)$ denotes the advantage function, and actually it corresponds to the temporal difference (TD) error.
Note that $A$ also depends on $\pi$, but it is omitted to simplify the notation.

With a policy regularization term $\Omega$~\citep{geist2019theory}, the original minimization target is reformulated as follows:
\begin{align}
    \mathcal{L}^\dagger(\pi) &= - \mathbb{E}_{a_t \sim \pi} \left [ A(s_t, a_t) \right ] + \mathbb{E}_{a_t \sim \pi} \left [ \Omega(s_t, a_t) \right ]
    \nonumber \\
    &= - \mathbb{E}_{a_t \sim b} \left [ \cfrac{\pi(a_t \mid s_t)}{b(a_t \mid s_t)} \left \{ A(s_t, a_t) - \Omega(s_t, a_t) \right \} \right ]
    \nonumber \\
    &= - \mathbb{E}_{a_t \sim b} \left [ \rho(s_t, a_t) A^\dagger(s_t, a_t) \right ]
    \label{eq:loss_reg_as} \\
    &= - \mathbb{E}_{a_t \sim b} \left [ \rho^\dagger(s_t, a_t) A(s_t, a_t) \right ]
    \label{eq:loss_reg_ds}
\end{align}
where a baseline policy $b(a_t \mid s_t)$ (e.g. the old version of $\pi$ or the one outputted from slowly updated target network~\citep{mnih2015human}) is introduced and a density ratio $\rho = \pi / b$ is derived in $\mathcal{L}^\dagger$.
This allows the agent to reuse past empirical data for training as experience replay~\citep{lin1992self,schaul2015prioritized}, where the actions have not been sampled from the current policy.
$A^\dagger(s_t, a_t)$ and $\rho^\dagger(s_t, a_t)$ are the surrogate versions defined as $A^\dagger(s_t, a_t) = A(s_t, a_t) - \Omega(s_t, a_t)$ and $\rho^\dagger(s_t, a_t) = \rho(s_t, a_t) (1 - \Omega(s_t, a_t) / A(s_t, a_t))$, respectively.

By applying a policy-gradient method to minimize $\mathcal{L}^\dagger(\pi)$ with Monte Carlo approximation, which eliminates the expectation operation, the policy parameterized by a parameters set $\theta$ (e.g. weights and biases in DNNs) can be optimized.
\begin{align}
    \nabla_\theta \mathcal{L}^\dagger(\pi) &\simeq - \nabla_\theta \{ \rho(s_t, a_t) A^\dagger(s_t, a_t) \}
    \nonumber \\
    &= - \rho(s_t, a_t) \tilde{A}^\dagger(s_t, a_t) \nabla_\theta \ln \pi(a_t \mid s_t)
    \label{eq:grad_reg} \\
    \theta &\gets \theta - \alpha \mathrm{SGD}(\nabla_\theta \mathcal{L}^\dagger(\pi))
    \label{eq:sgd}
\end{align}
where $\tilde{A}^\dagger = A^\dagger + \pi \nabla_\pi A^\dagger = A^\dagger + \rho \nabla_\rho A^\dagger$.
If $A^\dagger$ has no direct computational graph with $\pi$, $\tilde{A}^\dagger = A^\dagger$.
In general DRL, $\theta$ is updated using one of the stochastic gradient descent (SGD) optimizers like~\citep{ilboudo2020robust}.

\subsection{Proximal policy optimization}

In the clipping version of proximal policy optimization (PPO)~\citep{schulman2017proximal} (and its variant named PPO-RB~\citep{wang2020truly}), eq.~\eqref{eq:loss_reg_ds} is given as the following condition $\rho^\dagger = \rho^\mathrm{PPO}$, instead of the explicit $\Omega$, with a threshold parameter $\epsilon > 0$.
Note that, to simplify the rest of the description, the arguments (e.g. $(s_t, a_t)$) are omitted in most cases.
\begin{align}
    \rho^\mathrm{PPO} &= \begin{cases}
        -\eta \rho + (1 + \eta)(1 + \sigma\epsilon) & \sigma (\rho - 1) \geq \epsilon
        \\
        \rho & \mathrm{otherwise}
    \end{cases}
    \label{eq:ratio_ppo} \\
    \mathcal{L}^\mathrm{PPO}(\pi) &= - \mathbb{E}_{a_t \sim b} \left [ \rho^\mathrm{PPO}(s_t, a_t) A(s_t, a_t) \right ]
    \label{eq:loss_ppo}
\end{align}
where $\sigma$ denotes the sign of $A$.
$\eta \geq 0$ is for rollback to the baseline policy, which has been developed for PPO-RB, and only if $\eta = 0$, it reverts to the original PPO.

In this condition, $\Omega^\mathrm{PPO}$ for the specific regularization in PPO(-RB) can be analytically derived (see the appendix~\ref{app:omega_ppo}).
However, its heuristic design does not corresponds to divergence between two probabilities like KL divergence.

\subsection{Relative Pearson divergence}

\begin{figure*}[tb]
    \centering
    \subfigure[3D view]{
        \includegraphics[keepaspectratio=true,width=0.4\linewidth]{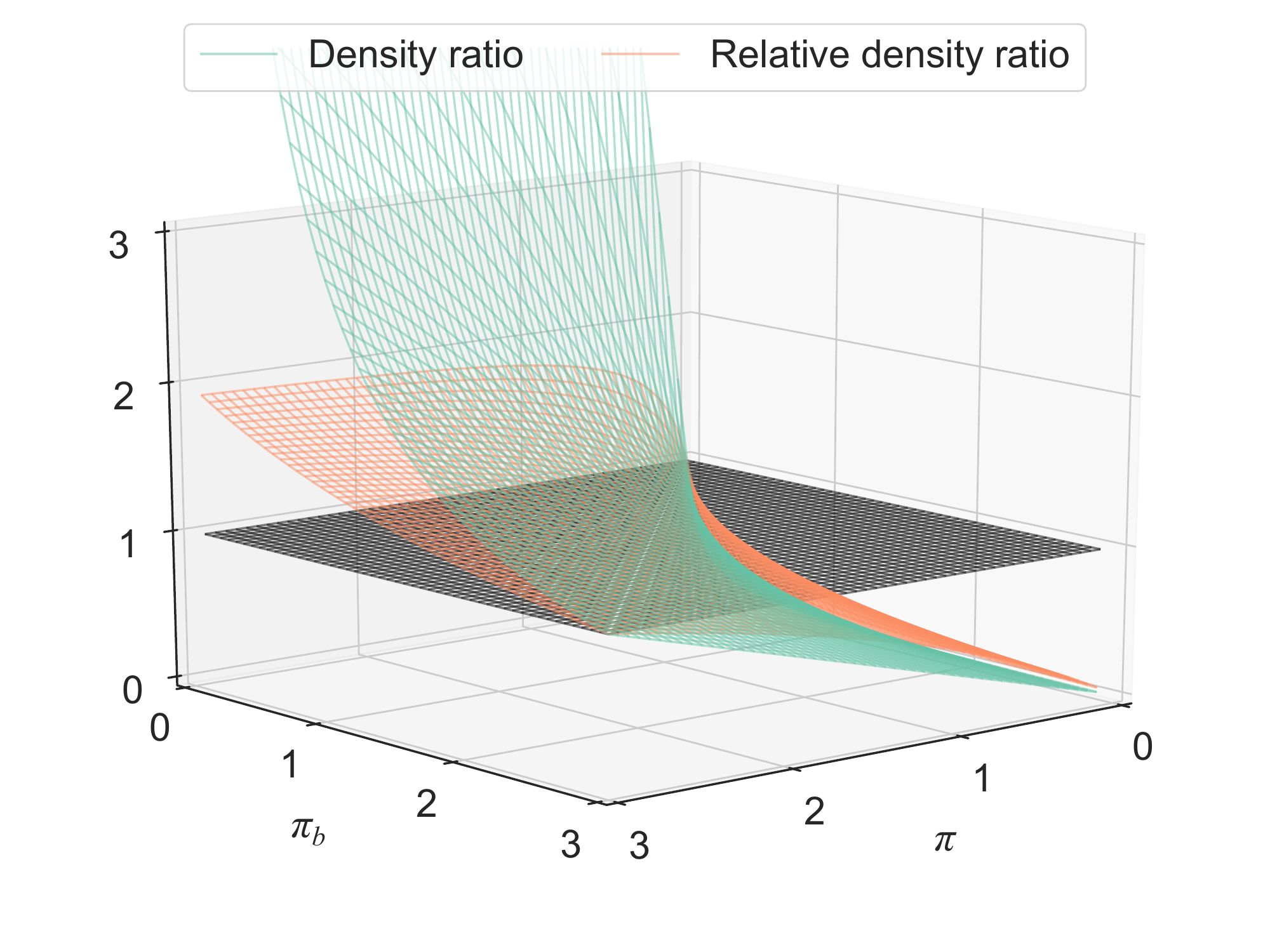}
    }
    \centering
    \subfigure[2D view]{
        \includegraphics[keepaspectratio=true,width=0.4\linewidth]{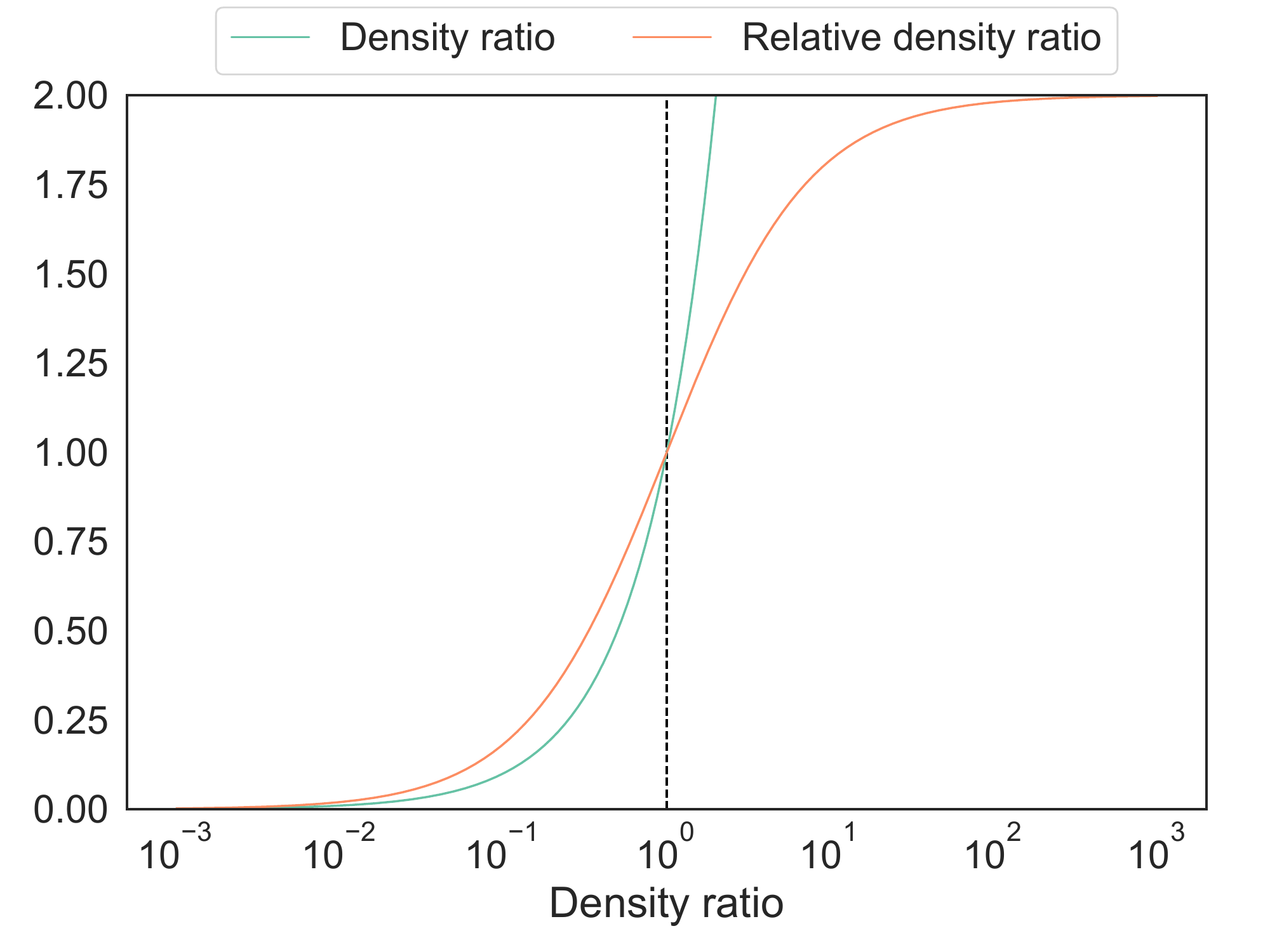}
    }
    \caption{Relative density ratio with $\beta = 0.5$:
        if and only if $\pi = b$, the both density ratios are equal to one;
        although the raw density ratio diverges as $b$ decreases and $\pi$ increases (i.e. $\rho$ increases), the relative density ratio has finite upper bounds $1/\beta$;
        in particular, if $\beta = 0.5$, the shape of relative density ratio becomes symmetric around one as shown in (b).
    }
    \label{fig:img_symmetric}
\end{figure*}

To inherit the abilities of PPO (eqs.~\eqref{eq:ratio_ppo} and~\eqref{eq:loss_ppo}) and overcome its issues, the following conditions are considered.
\begin{enumerate}
    \item When $\rho = 1$, the term inside the expectation of the loss function should be $\rho A$.
    \item Only when $\rho$ is on the threshold, the gradient must be zero and there is only one local maximum.
    \item For the explicit constraint between $\pi$ and $b$, one of the f-divergences, which is the generalized divergence~\citep{liese2006divergences}, is desired to be utilized.
    \item Symmetrical shape of $\rho^\dagger$ is desired to symmetrically clip it and to adjust the threshold $\epsilon$ easily according to the error scale of $\rho^\dagger$ from its center.
\end{enumerate}

KL divergence, which is utilized in the prior version of PPO~\citep{schulman2017proximal}, cannot satisfy all of them (especially, the symmetry).
Thus, the alternative divergence among one of the f-divergences that satisfies all the above is desired to be selected even heuristically.
This study finds that relative Pearson (RPE) divergence~\citep{yamada2013relative,sugiyama2013direct} has the potential to satisfy them.

Specifically, the standard Pearson (PE) divergence between two policies, $\pi$ and $b$, is given at first.
\begin{align}
    \mathrm{PE}(\pi, b) = \mathbb{E}_{a \sim b} \left [ \cfrac{1}{2}\left ( \rho(a) - 1 \right )^2 \right ]
    \label{eq:div_pe}
\end{align}
Due to $\int b(a \mid s) \rho(s, a) da = \int \pi(a \mid s) da = 1$, the mean of $\rho$ is equal to $1$.
That is, this PE divergence means the expected squared error of $\rho$ from its mean.
This is non-negative and vanishes if and only if $\pi = b$.

Instead of $b$, RPE divergence introduces the relative density function with a mixture ratio $\beta \in [0, 1]$ as follows:
\begin{align}
    \pi_\beta = \beta \pi + (1 - \beta) b
    \label{eq:pi_beta}
\end{align}
In addition, the relative density ratio $\rho_\beta$ is defined using $\pi_\beta$.
\begin{align}
    \rho_\beta = \cfrac{\pi}{\pi_\beta} = \cfrac{\rho}{\beta \rho + 1 - \beta}
    \label{eq:rho_beta}
\end{align}
where the term on the right side can be derived by multiplying it with $1 = b^{-1}/b^{-1}$.
As important properties, $\rho_\beta$ is finite within $[0, 1/\beta)$, and its mean is still $1$.
Therefore, if $\beta = 0.5$, $\rho_\beta$ obtains the symmetry, as shown in Fig.~\ref{fig:img_symmetric}.

By replacing $\rho$ in eq.~\eqref{eq:div_pe} to $\rho_\beta$, RPE divergence is defined as follows:
\begin{align}
    \mathrm{PE}_\beta(\pi, b) &= \mathbb{E}_{a \sim \pi_\beta} \left [ \cfrac{1}{2}\left ( \rho_\beta(a) - 1 \right )^2 \right ]
    \nonumber \\
    &= \mathbb{E}_{a \sim \pi_\beta} \left [ \cfrac{1}{2}\left ( \cfrac{\rho(a)}{\beta \rho(a) + (1 - \beta)} - 1 \right )^2 \right ]
    \label{eq:div_rpe}
\end{align}

\section{Proposed method}

\subsection{Proximal policy optimization with relative Pearson divergence: PPO-RPE}

\begin{algorithm}[tb]
    \caption{PPO-RPE: the value function $Q$ and/or $V$ will be trained in parallel according to Bellman equation.}
    \label{alg:pporpe}
    \begin{algorithmic}[1]
        \State{Given environment as $p_e(s_{t+1} \mid s_t, a_t)$ and $r(s_t, a_t)$}
        \State{Set $\beta \in [0, 1]$ (0.5 is the default value for symmetry)}
        \State{Set $\gamma \in [0, 1)$ (0.99 is the default value)}
        \State{Initialize the policy $\pi$ with $\theta$ and the baseline policy $b$}
        \State{Initialize optimizer as SGD with learning rate $\alpha$}
        \While{True}
            \\\hrulefill\Comment{Interaction with environment}
            \State{$s_t \sim p_e(s_t \mid s_{t-1}, a_{t-1})$}
            \State{$a_t \sim b(a_t \mid s_t)$}
            \State{$s_{t+1} \sim p_e(s_{t+1} \mid s_t, a_t)$}
            \State{$r_t = r(s_t, a_t)$}
            \\\hrulefill\Comment{Main formulas of PPO-RPE}
            \State{$A = Q(s_t, a_t) - V(s_t) \simeq r_t + \gamma V(s_{t+1}) - V(s_t)$}
            \State{$\rho = \cfrac{\pi(a_t \mid s_t)}{b(a_t \mid s_t)}$}
            \State{$\rho_\beta = \cfrac{\rho}{1 - \beta + \beta \rho}$}
            \State{Get $C$ from Alg.~\ref{alg:threshold} given $A$ and $\rho_\beta$ as inputs}
            \State{$\tilde{A}^\mathrm{RPE} = A - C (\rho_\beta - 1)\left \{ \beta (\rho_\beta  - 1) + 2 (1 - \beta) \cfrac{\rho_\beta}{\rho} \right \}$}
            \\\hrulefill\Comment{Update of policies}
            \State{$g = - \rho \tilde{A}^\mathrm{RPE} \nabla_\theta \ln \pi(a_t \mid s_t)$}
            \State{$\theta \gets \theta - \alpha \mathrm{SGD}(g)$}
            \State{Update $b$ according to $\pi$ with $\theta$}
            \State{$t \gets t + 1$}
        \EndWhile
    \end{algorithmic}
\end{algorithm}

This paper proposes a proximal policy optimization with relative Pearson divergence, so-called PPO-RPE.
The explicit f-divergence minimization can softly constrain the latest policy $\pi$ to the baseline one $b$.
In particular, by employing RPE divergence, the gain of this regularization $C$ can be adaptively tuned as PPO-like (but adaptive, not fixed) threshold, as introduced in the next section.
The main process of PPO-RPE except the adaptive threshold for determining the gain $C$ is summarized in Alg.~\ref{alg:pporpe}.

\subsubsection{Minimization target}

To make PPO-RPE be a sub-class of policy-regularized RL described in eq.~\eqref{eq:loss_reg_as}, the regularization term for PPO-RPE, $\Omega^\mathrm{RPE}$, is derived.
With $\pi_\beta$ in eq.~\eqref{eq:pi_beta} and $\rho_\beta$ in eq.~\eqref{eq:rho_beta}, $\Omega^\mathrm{RPE}$ is given as follows:
\begin{align}
    \Omega^\mathrm{RPE} &= C \cfrac{(\rho_\beta - 1)^2}{\rho_\beta}
    = C \cfrac{1 - \beta + \beta \rho}{\rho} (\rho_\beta - 1)^2
    \label{eq:reg_rpe}
\end{align}
where $C$ again denotes the gain of this regularization.
The expectation w.r.t $\pi$ of this regularization is equivalent to eq.~\eqref{eq:div_rpe} amplified by the fixed $C$ as shown in below.
\begin{align}
    \mathbb{E}_{a \sim \pi}[\Omega^\mathrm{RPE}] &= \int \pi C (\rho_\beta - 1)^2 \cfrac{\pi_\beta}{\pi} da
    \nonumber \\
    &= \int \pi_\beta C (\rho_\beta - 1)^2 da
    \nonumber \\
    &= C \mathrm{PE}_\beta(\pi, b)
\end{align}
Note that if $C$ depends on $\pi_\beta$ (or $\pi$), this derivation is not exact but sufficiently relevant.

By substituting eq.~\eqref{eq:reg_rpe} for eq.~\eqref{eq:loss_reg_as}, the following regularized loss function is derived for PPO-RPE.
\begin{align}
    \mathcal{L}^\mathrm{RPE}(\pi) &= - \mathbb{E}_{a_t \sim b} \left [ \rho A - C (1 - \beta + \beta \rho) (\rho_\beta - 1)^2 \right ]
    \nonumber \\
    &= - \mathbb{E}_{a_t \sim b} \left [ \rho A^\mathrm{RPE} \right ]
    \label{eq:loss_rpe}
\end{align}
where $\Omega^\mathrm{RPE}$ is included in $A^\mathrm{RPE}$, and it can be represented as $\rho^\mathrm{RPE} A$ with $\rho^\mathrm{RPE} = \rho A^\mathrm{RPE} / A$.

\subsubsection{Policy gradient}

Here, the policy gradient for eq.~\eqref{eq:loss_rpe} is analytically derived.
According to eq.~\eqref{eq:grad_reg}, $\tilde{A}^\mathrm{RPE} = A^\mathrm{RPE} + \rho \nabla_\rho A^\mathrm{RPE}$ is computed analytically as follows (its derivation is in the appendix~\ref{app:grad_rpe}):
\begin{align}
    \tilde{A}^\mathrm{RPE} &= A \left [ 1 - \cfrac{C}{A}(\rho_\beta - 1)\left \{ \beta (\rho_\beta  - 1) + 2 (1 - \beta) \cfrac{\rho_\beta}{\rho} \right \} \right ]
    \label{eq:grad_rpe}
\end{align}
This is substituted for eq.~\eqref{eq:grad_reg}, and used for updating the parameters set $\theta$ in eq.~\eqref{eq:sgd}.

The second term in the square bracket is consistent with the regularization, and vanishes if and only if $\rho = \rho_\beta = 1$ (i.e. $\pi = b$) regardless of $C$ and $\beta$.
Namely, with this policy gradient, $\pi$ has the potential to converge on $\pi^*$, which maximizes $A$ (more specifically, the action-value function $Q$), while being regularized to $b$ during the policy updates.

\subsection{Adaptive threshold for designing gain of regularization}

\begin{algorithm}[tb]
    \caption{Adaptive threshold design for PPO-RPE}
    \label{alg:threshold}
    \begin{algorithmic}[1]
        \State{Set $\beta \in [0, 1]$ (0.5 is the default value)}
        \State{Set $\lambda \in (0, 1)$ (0.999 is the recommended value)}
        \State{Set $\kappa \in (0, 1)$ (0.5 is the recommended value)}
        \State{Set $\underline{\Delta} \in (0, 0.5)$ (0.1 is the recommended value)}
        \State{Initialize $\Delta_\mathrm{max} = 0$ (i.e. no prior information)}
        \State{Initialize $\Delta = 1$ (i.e. theoretical maximum value)}
        \While{True}
            \\\hrulefill\Comment{Computation of $C$}
            \State{Get $A$ and $\rho_\beta$ from Alg.~\ref{alg:pporpe}}
            \State{$\sigma = \mathrm{sign}(A)$}
            \State{$\epsilon = \kappa \max(\min(\Delta, 1 - \underline{\Delta}), \underline{\Delta})$}
            \State{$C = \cfrac{|A|}{\beta \sigma \epsilon^2 + 2 \epsilon \{1 - \beta (1 + \sigma \epsilon)\}}$}
            \State{Yield $C$ to Alg.~\ref{alg:pporpe}}
            \\\hrulefill\Comment{Update of $\Delta$}
            \State{$\Delta_\mathrm{max} \gets \max(\lambda \Delta_\mathrm{max}, |\rho_\beta - 1|)$}
            \State{$\Delta \gets \lambda \Delta + (1 - \lambda) \Delta_\mathrm{max}$}
        \EndWhile
    \end{algorithmic}
\end{algorithm}

Actually, the gain for regularization, $C$, is hard to be tuned since the ratio between $A$ for the original purpose and $\Omega^\mathrm{RPE}$ for the regularization is task- and algorithm-dependent.
In contrast to $C$, PPO utilizes the threshold with $\epsilon$ for tuning the strength of regularization.
The threshold allows for more intuitive design, and in fact, the original paper~\citep{schulman2017proximal} shows the recommended range of $\epsilon$ for any task.

Inspired by this, $C$ is converted to the PPO-like threshold, which releases from the task dependency.
In addition, by applying $\epsilon$ to the relative density ratio $\rho_\beta$, not the raw density ratio $\rho$, the symmetric threshold works in the same symmetric domain.
Such symmetry enables PPO-RPE to easily estimate the algorithm-dependent error scale of $\rho_\beta$ from its center $1$, which would be utilized for the adaptive design of $\epsilon$.
The pseudo code for the adaptive threshold is described in Alg.~\ref{alg:threshold}.

\subsubsection{Conversion from gain to threshold}

We focus on the fact that the density ratio $\rho$ is asymmetric around $1$, and not consistent with the symmetric threshold in PPO, causing unbalanced regularization.
Instead, PPO-RPE can serve the threshold for the relative density ratio $\rho_\beta$ with $\beta = 0.5$, which is in the symmetric domain (see again Fig.~\ref{fig:img_symmetric}).
Given $\epsilon \in (0, 1)$ (will be adapted later), the following condition is considered.
\begin{align}
    \rho_\beta^\epsilon &= 1 + \sigma \epsilon
    \label{eq:thr_rb} \\
    \rho^\epsilon &= \cfrac{(1 - \beta) (1 + \sigma \epsilon)}{1 - \beta (1 + \sigma \epsilon)}
    = 1 + \cfrac{\sigma \epsilon}{1 - \beta (1 + \sigma \epsilon)}
    \label{eq:thr_r}
\end{align}
where $\rho^\epsilon$ can be inversely derived from eq.~\eqref{eq:rho_beta}.
Note that $\sigma \epsilon / (1 - \beta (1 + \sigma \epsilon))$ in $\rho^\epsilon$ yields the asymmetric threshold for $\rho$ domain, and if $\beta = 0$, it is reverted to be symmetric.

Only at $\rho_\beta^\epsilon$ (and $\rho^\epsilon$), in order to softly constrain the latest policy $\pi$ to the baseline $b$, $\tilde{A}^\mathrm{RPE}$ is desired to be zero, which makes $\mathcal{L}^\mathrm{RPE}$ be convex.
Using this condition, $C$ can be derived from eq.~\eqref{eq:grad_rpe} as follows:
\begin{align}
    0 &= 1 - \cfrac{C}{A} (\rho_\beta^\epsilon - 1) \left \{ \beta (\rho_\beta^\epsilon  - 1) + 2 (1 - \beta) \cfrac{\rho_\beta^\epsilon}{\rho^\epsilon} \right \}
    \nonumber \\
    &= 1 - \cfrac{C}{A} \sigma \epsilon \left \{ \beta \sigma \epsilon + 2 (1 - \beta) \cfrac{1 - \beta (1 + \sigma \epsilon)}{1 - \beta} \right \}
    \nonumber \\
    C &= \cfrac{|A|}{\beta \sigma \epsilon^2 + 2 \epsilon \{1 - \beta (1 + \sigma \epsilon)\}}
    \label{eq:gain_rpe}
\end{align}
where $A / \sigma = |A|$ is utilized.
Indeed, $C$ designed above is regarded as the function of $A$, and therefore, yields the adaptability to various tasks with different scales of reward function.

By substituting eq.~\eqref{eq:gain_rpe} for eq.~\eqref{eq:grad_rpe}, PPO-RPE achieves the following policy gradient term with $\epsilon$ instead of $C$.
\begin{align}
    \tilde{A}^\mathrm{RPE} = A - \cfrac{|A| (\rho_\beta - 1) \left \{\beta (\rho_\beta - 1) - 2 (1 - \beta) \cfrac{\rho_\beta}{\rho} \right \}}{\beta \sigma \epsilon^2 - 2 \epsilon\left \{1 - \beta (1 + \sigma \epsilon) \right \}}
    \label{eq:grad_rpe_thre}
\end{align}
Using this, the parameters set $\theta$ is updated according to SGD with the policy gradient $- \rho \tilde{A}^\mathrm{RPE} \nabla_\theta \ln \pi(a_t \mid s_t)$.

\begin{figure}[tb]
    \centering
    \includegraphics[keepaspectratio=true,width=0.95\linewidth]{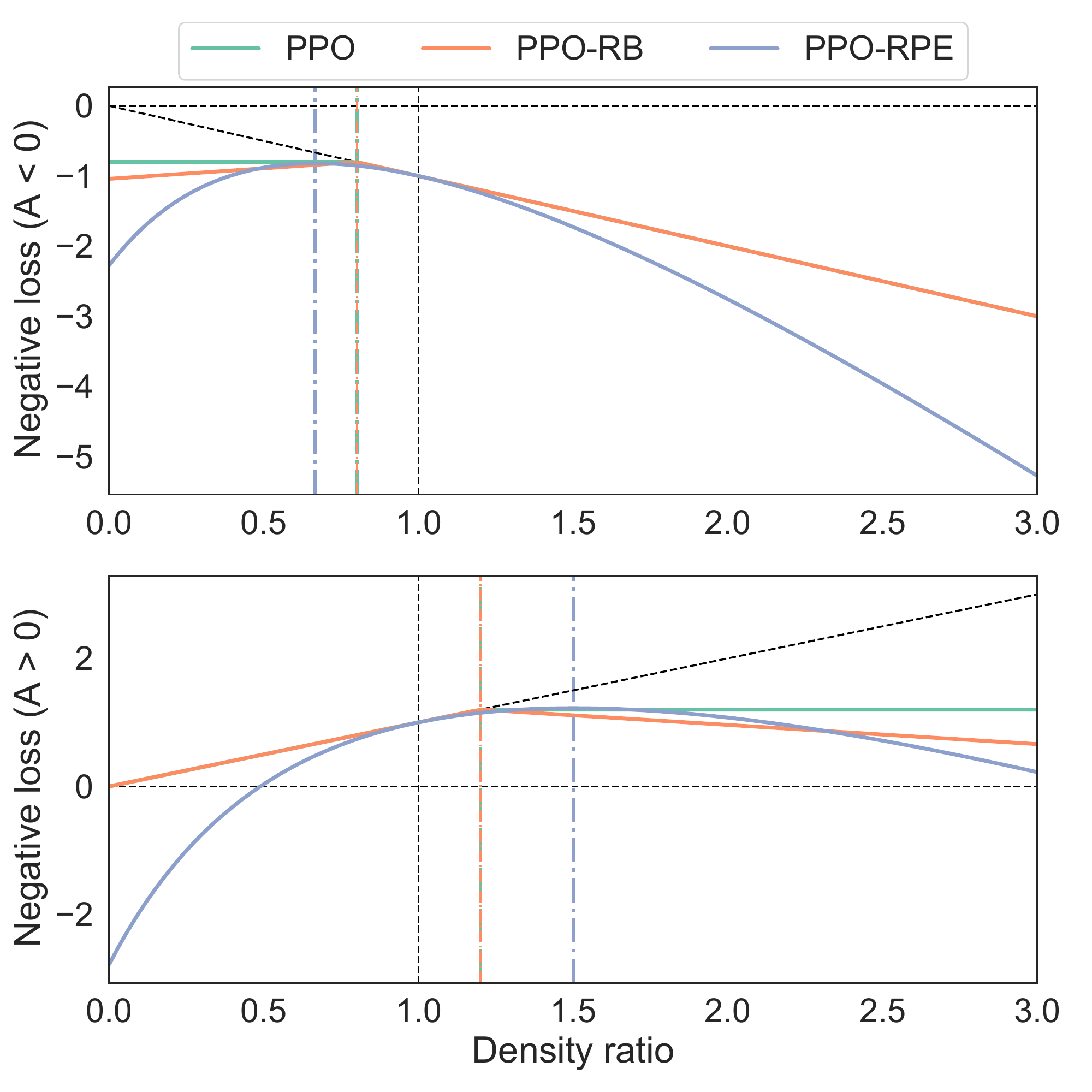}
    \caption{Negative loss functions for the respective methods:
        dash-dot vertical lines with colors corresponding to the legends indicate the respective thresholds;
        PPO and PPO-RB place the two thresholds (when $A > 0$ and $A < 0$, respectively) symmetrically around one;
        in contrast, PPO-RPE sets its two local maximum values as thresholds (when $A > 0$ and $A < 0$, respectively) asymmetrically in the domain of the raw density ratio;
        in addition, PPO-RPE has the capability to softly constrain the latest policy to the baseline policy.
    }
    \label{fig:img_loss}
\end{figure}

Here, all the examples of the negative loss functions (i.e. $\rho A^\dagger$ or $\rho^\dagger A$) for the proposed and conventional methods are illustrated in Fig.~\ref{fig:img_loss}.
As can be seen in this figure, PPO-RPE has the asymmetric threshold in $\rho$ domain, which is symmetric in $\rho_{\beta=0.5}$ domain.

\subsubsection{Adaptive threshold using estimated error scale}

In eq.~\eqref{eq:grad_rpe_thre}, we can see that the threshold-based gain for regularization can adapt to differences in reward scales for different tasks.
However, the mean of divergence between $\pi$ and $b$ depends on the learning algorithm.
For example, when $b$ is given by the target network, when experience replay~\citep{lin1992self} is not used, or even when the learning rate $\alpha$ is changed, the mean would vary.
The algorithm-dependent design of threshold is therefore required, but as can be imagined, hand-tuning is burdensome since the recent DRL methods are constructed like by patchworks of various modules and there are many combinations.
In addition, it is difficult to completely eliminate task dependency only by the threshold-based design.
From the earlier work~\citep{kobayashi2021proximal}, this paper additionally contributes to design the adaptive threshold $\epsilon$.

Specifically, thanks to the symmetry of $\rho_\beta$ (with $\beta = 0.5$) and its consistency with the threshold, its maximum error scale from its center (i.e. one) is revealed as one: $|0 - 1| = |1/0.5 - 1| = 1$.
With this fact, $\epsilon$ can be interpreted as defining a threshold value for the algorithm-independent (i.e. absolute) maximum error scale, although the algorithm-dependent (i.e. relative) maximum error scale may vary from it.
It is therefore expected that if $\epsilon$ is given for the relative maximum error scale, it would release from the algorithm dependency.

To estimate the relative maximum error scale in the recent past, $\Delta$, the following heuristics is implemented with three hyperparameters: $\lambda \in (0, 1)$ for smoothing updates; $\kappa \in (0, 1)$ for the relative threshold; and $\underline{\Delta} \in (0, 0.5)$ and $\overline{\Delta} = 1 - \underline{\Delta}$ for numerical stability;.
\begin{align}
    \Delta_\mathrm{max} &\gets \max(\lambda \Delta_\mathrm{max}, |\rho_\beta - 1|)
    \label{eq:err_max} \\
    \Delta &\gets \lambda \Delta + (1 - \lambda) \Delta_\mathrm{max}
    \label{eq:err_scale} \\
    \epsilon &= \kappa \max(\min(\Delta, \overline{\Delta}), \underline{\Delta})
    \label{eq:thr_adapt}
\end{align}
Here, $\Delta_\mathrm{max}$ stores the recent maximum value, then it is smoothly transferred into $\Delta$.
Since max operator switches the output value discretely, this two-step update provides smooth estimation of $\Delta$ and suppresses regularization fluctuations.
By giving $\epsilon$ in this way, the adaptive threshold in PPO-RPE is obtained against the relative maximum error scale that depends on the learning algorithm, easing algorithm-specific tuning.

As a remark, although the new hyperparameters have been introduced here, they can be given regardless of other implementations:
for $\lambda$, 0.999, which is also used in the most of SGDs~\citep{ziyin2020laprop,kobayashi2021towards,ilboudo2020robust} for relatively slow updates, would be the appropriate value;
$\kappa$ can be fixed to place the local maximum of the regularization onto half of the relative maximum error scale, i.e. 0.5;
and $\underline{\Delta}$ can be designed based on the minimum recommended value of $\epsilon$ in PPO, i.e. 0.1.

\section{Simulations}

\subsection{Implementations}

\begin{table}[tb]
    \caption{Common hyperparameters for the simulations}
    \label{tab:parameter}
    \centering
    \begin{tabular}{ccc}
        \hline\hline
        Symbol & Meaning & Value
        \\
        \hline
        $N$ & Number of neurons & 100
        \\
        $L$ & Number of layers & 5
        \\
        $\gamma$ & Discount factor & 0.99
        \\
        $\alpha$ & Learning rate & $3 \times 10^{-4}$
        \\
        $(\tau, \nu)$ & For t-soft update~\citep{kobayashi2021t} & (0.5, 1.0)
        \\
        $\beta_{DE}$ & Gain for entropy regularization & 0.01
        \\
        $\beta_{TD}$ & Gain for TD regularization & 0.01
        \\
        $(N_c, N_e, N_b)$ & For experience replay & $(10^5, 10^2, 10^2)$
        \\
        $(\lambda_\mathrm{max}^1, \lambda_\mathrm{max}^2, \kappa)$ & For adaptive eligibility traces~\citep{kobayashi2022adaptive} & (0.5, 0.95, 10)
        \\
        \hline\hline
    \end{tabular}
\end{table}

\begin{figure}[tb]
    \centering
    \includegraphics[keepaspectratio=true,width=0.9\linewidth]{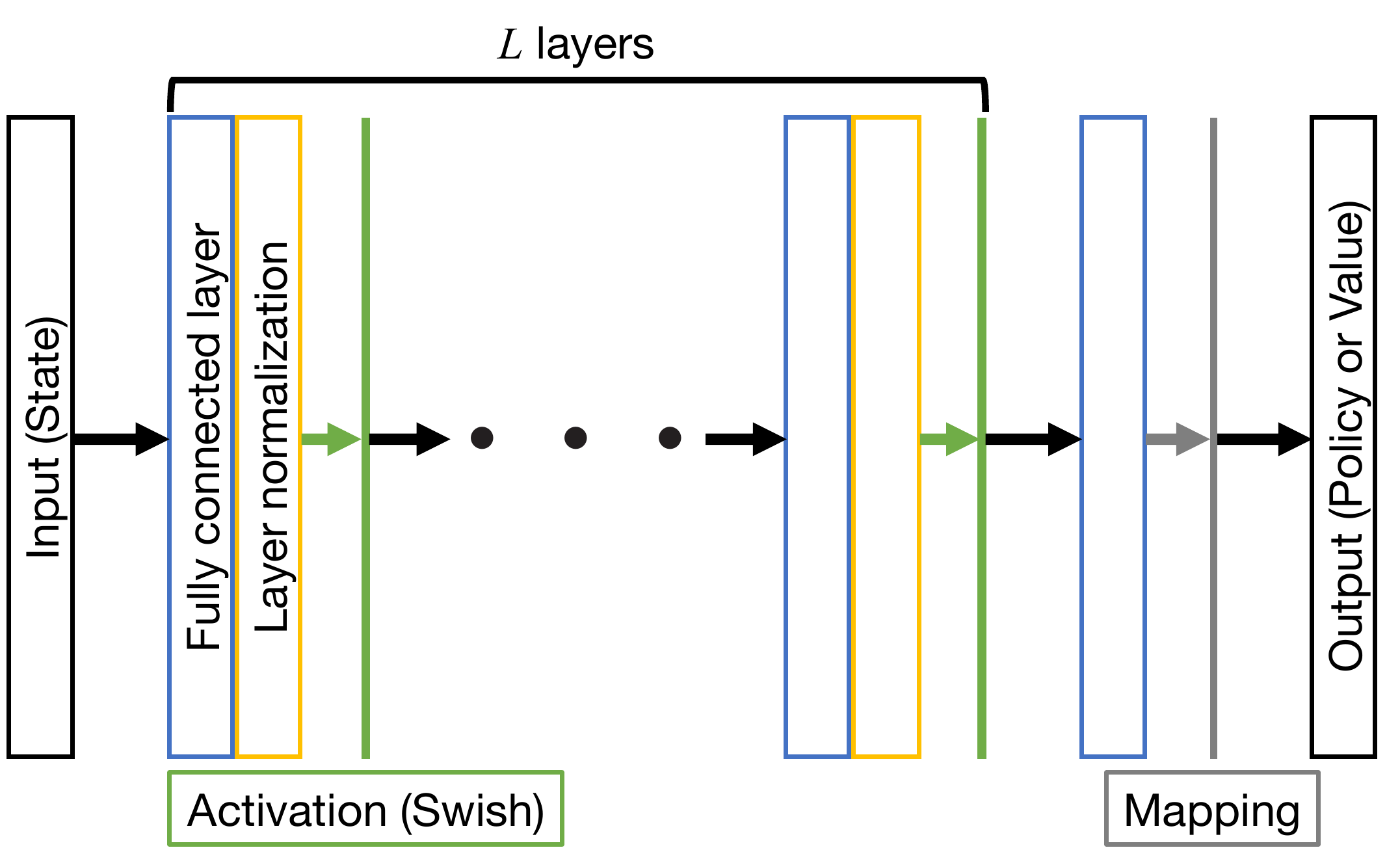}
    \caption{Network architecture in actor for policy or critic for value function:
        state is inputted to $L$ series-connected modules;
        each module contains a fully connected layer, a layer normalization~\citep{ba2016layer}, and a swish activation function~\citep{elfwing2018sigmoid};
        an additional fully connected layer uses for shaping the features given by $L$ modules to the output.
    }
    \label{fig:img_network}
\end{figure}

In this paper, the proposed method with its hyperparameters is implemented in almost the same way as in~\citep{kobayashi2021t}.
The details are introduced below.
Note that all the hyperparameters commonly used are listed in Table~\ref{tab:parameter}.

First, the policy $\pi$ is parameterized by student-t distribution~\citep{kobayashi2019student} with neural networks.
$d_a$-dimensional student-t distribution has three model parameters: $\mu \in \mathbb{R}^{d_a}$; $\sigma \in \mathbb{R}^{d_a}_+$; and $\nu \in \mathbb{R}_+$.
They are given from the networks with $d_s$-dimensional state space inputs.
The value function $V$ is approximated by the corresponding networks with the same architecture as that of $\pi$.

The networks contains $L=5$ fully connected layers with $N=100$ neurons and pairs of layer normalization~\citep{ba2016layer} and Swish activation function~\citep{elfwing2018sigmoid} for nonlinearity (also see Fig.~\ref{fig:img_network}).
These implementations are built on PyTorch~\citep{paszke2017automatic}.
As an optimizer of the networks, a robust SGD, i.e., LaProp~\citep{ziyin2020laprop} with t-momentum~\citep{ilboudo2020robust} and d-AmsGrad~\citep{kobayashi2021towards} (so-called td-AmsProp), is employed with their default parameters except the learning rate.

To learn the networks stably, a target network with t-soft update~\citep{kobayashi2021t} is employed for stable and efficient improvement.
Here, the baseline policy $b$ is regarded as the output from the target network for $\pi$, that is, $b$ is a kind of the old policy.
In addition, the policy entropy regularization based on SAC~\citep{haarnoja2018soft} with a regularization weight $\beta_{DE}$; and the TD regularization~\citep{parisi2019td} with a regularization weight $\beta_{TD}$ are combined.

For acceleration of learning, either or both of adaptive eligibility traces~\citep{kobayashi2022adaptive} and the experience replay are employed.
Although the adaptive eligibility traces are with the same hyperparameter as~\citep{kobayashi2021t,kobayashi2021proximal}, the buffer size of the experience replay is reduced due to memory limitation.
According to them, the proposed method will appropriately modify the threshold $\epsilon$ adaptively.

\subsection{Environments}

\begin{table*}[tb]
    \caption{Environments simulated by Pybullet Gym~\citep{brockman2016openai,coumans2016pybullet}}
    \label{tab:environment}
    \centering
    \begin{tabular}{ccccc}
        \hline\hline
        ID & Name & State space $d_s$ & Action space $d_a$ & Episode $E$
        \\ \hline
        InvertedPendulumBulletEnv-v0 & InvertedPendulum & 5 & 1 & 200
        \\
        InvertedPendulumSwingupBulletEnv-v0 & Swingup & 5 & 1 & 200
        \\
        InvertedDoublePendulumBulletEnv-v0 & DoublePendulum & 9 & 1 & 2000
        \\
        HopperBulletEnv-v0 & Hopper & 15 & 3 & 2000
        \\
        Walker2DBulletEnv-v0 & Walker2D & 22 & 6 & 2000
        \\
        HalfCheetahBulletEnv-v0 & HalfCheetah & 26 & 6 & 2000
        \\
        AntBulletEnv-v0 & Ant & 28 & 8 & 2000
        \\ \hline\hline
    \end{tabular}
\end{table*}

In this experiment, seven benchmark tasks are simulated on Pybullet dynamical engine~\citep{coumans2016pybullet} wrapped by OpenAI Gym~\citep{brockman2016openai} (see Table~\ref{tab:environment}).
They can be divided into three simple and four complex tasks.
That is, the first three tasks (i.e. InvertedPendulum, Swingup, and DoublePendulum) are the simple ones with only one-dimensional action space, although they are nonlinear systems.
The remaining tasks (i.e. Hopper, Walker2D, HalfCheetah, and Ant) are the complex locomotion control tasks with multivariate action space and more than 10-dimensional state space.
Totally, 20 trials for the simple tasks and 10 trials for the complex tasks with each condition are conducted with different random seeds.
Note that the robot's power in Walker2D was insufficient to walk, hence, it is amplified by a factor of 3.75 (i.e. 0.4 to 1.5).

\begin{figure*}[tb]
    \centering
    \subfigure[InvertedPendulum w/ eligibility traces]{
        \includegraphics[keepaspectratio=true,width=0.315\linewidth]{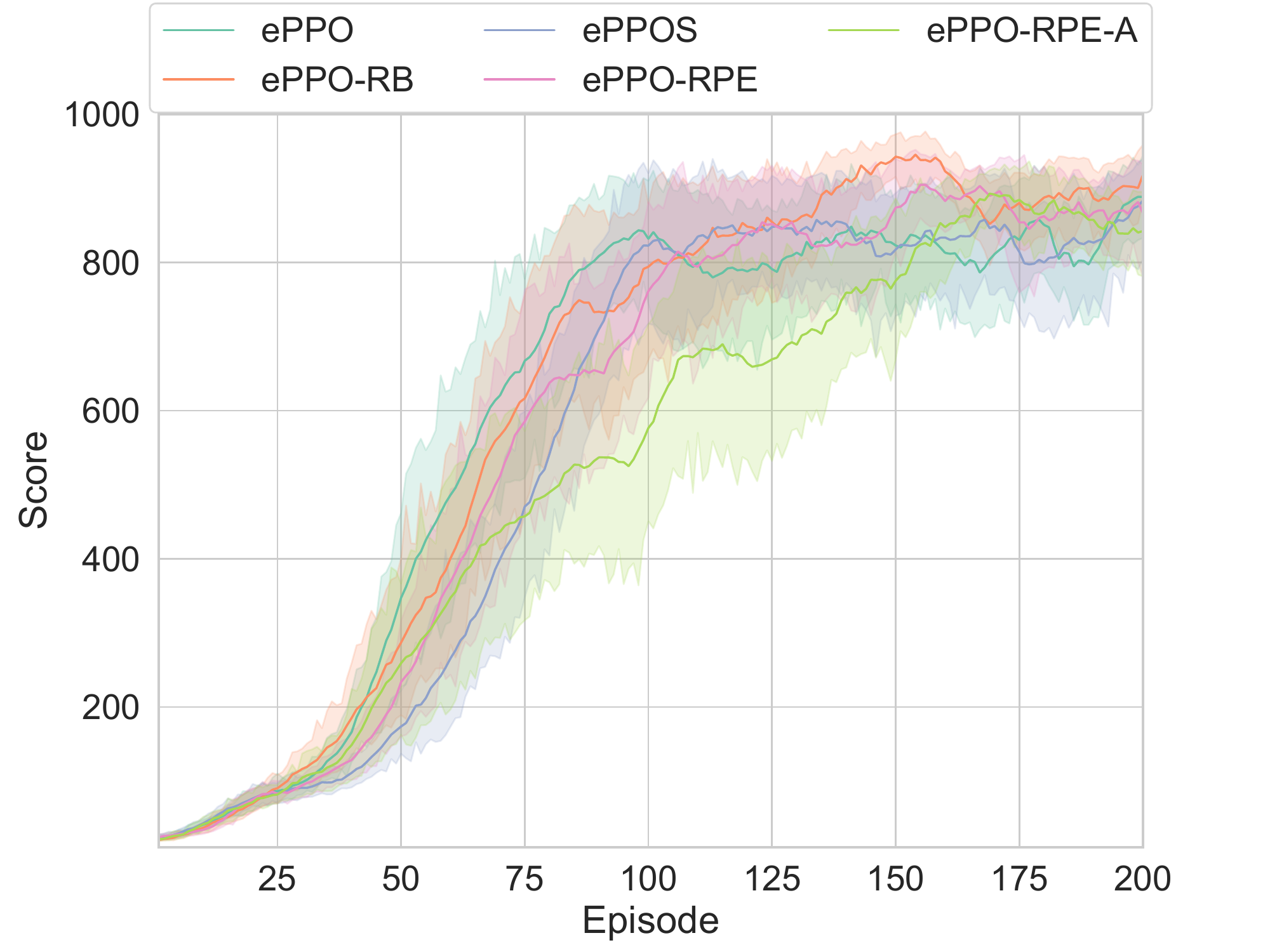}
    }
    \centering
    \subfigure[Swingup w/ eligibility traces]{
        \includegraphics[keepaspectratio=true,width=0.315\linewidth]{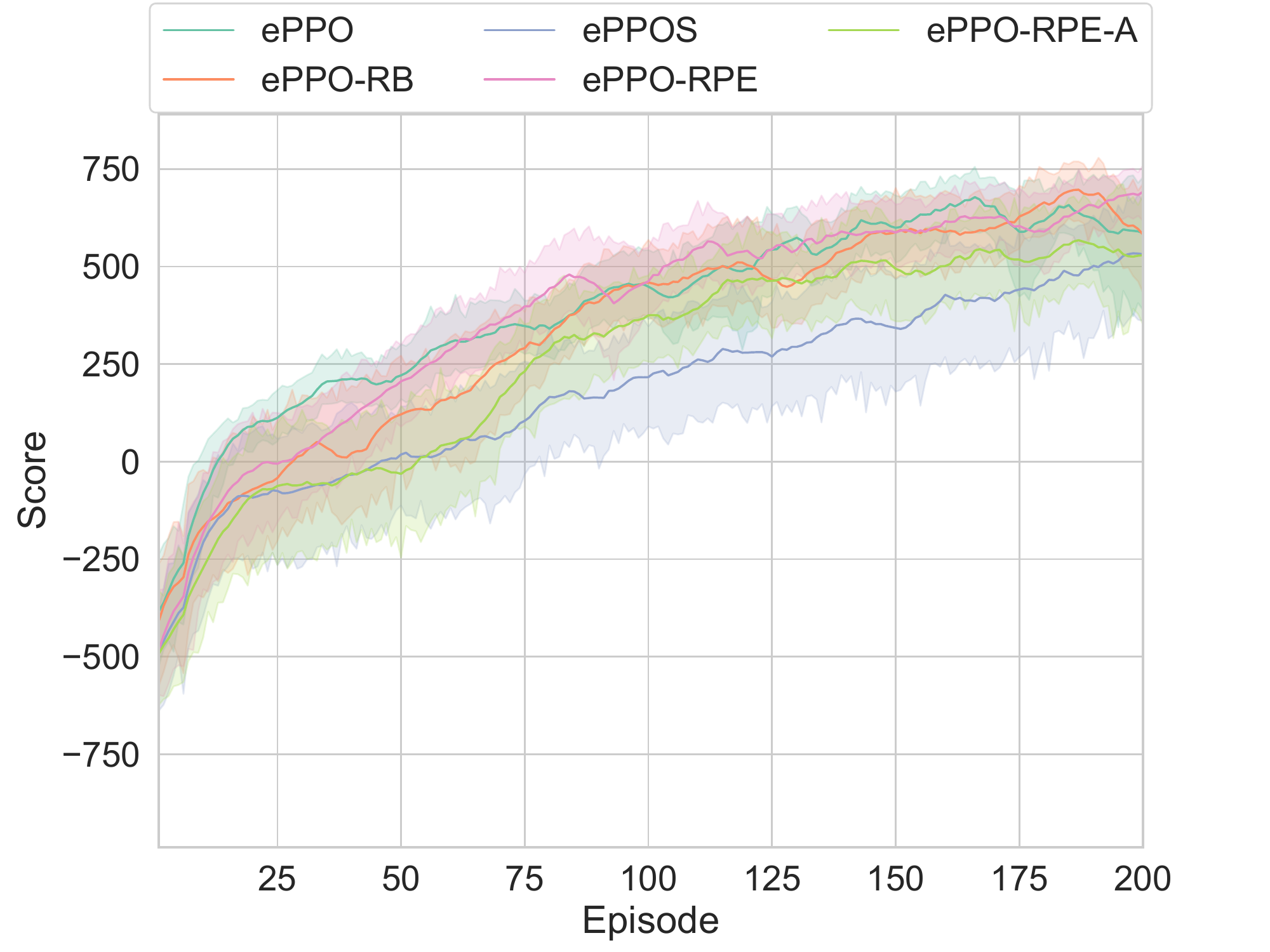}
    }
    \centering
    \subfigure[DoublePendulum w/ eligibility traces]{
        \includegraphics[keepaspectratio=true,width=0.315\linewidth]{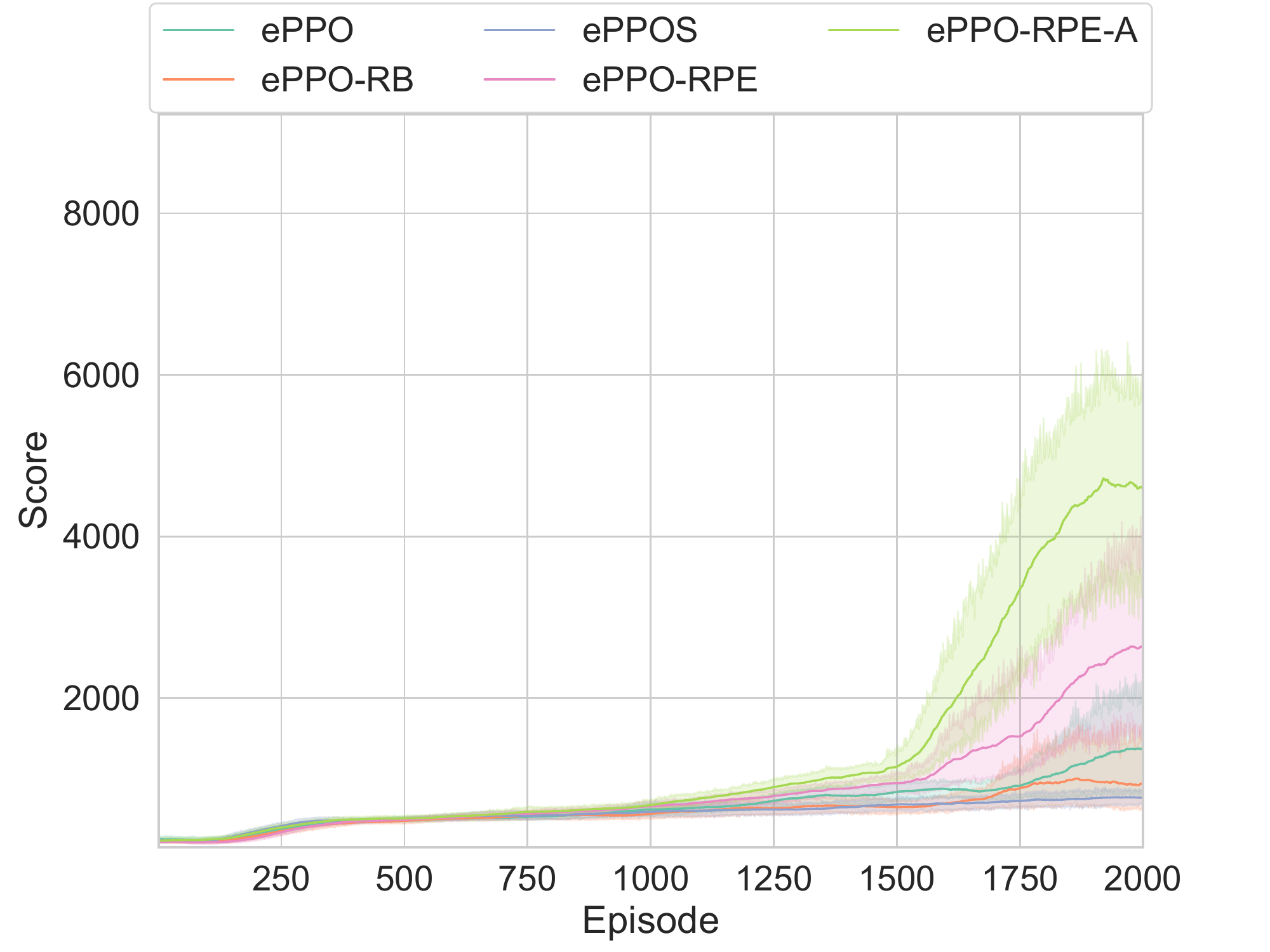}
    }
    \centering
    \subfigure[InvertedPendulum w/ experience replay]{
        \includegraphics[keepaspectratio=true,width=0.315\linewidth]{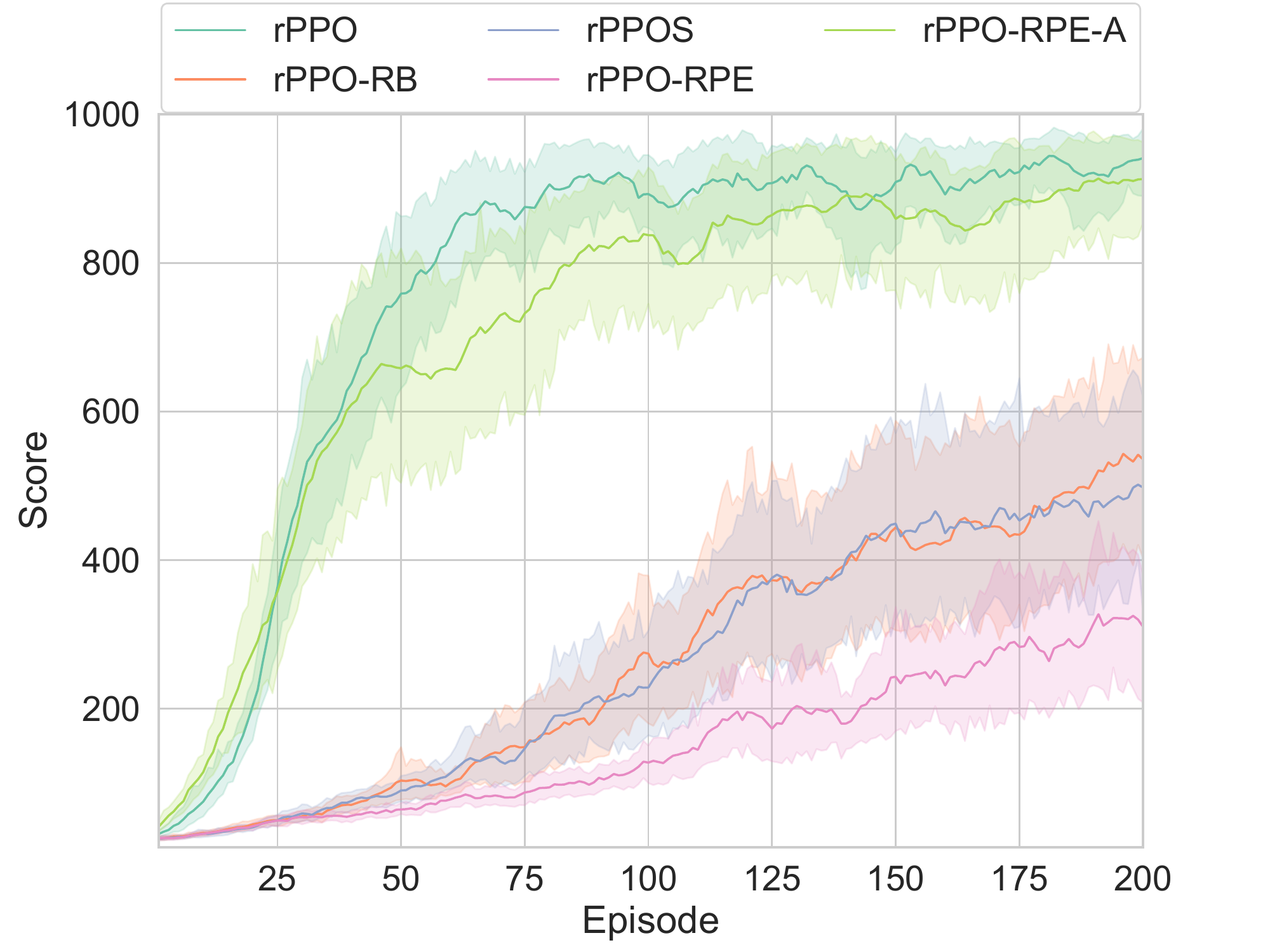}
    }
    \centering
    \subfigure[Swing up w/ experience replay]{
        \includegraphics[keepaspectratio=true,width=0.315\linewidth]{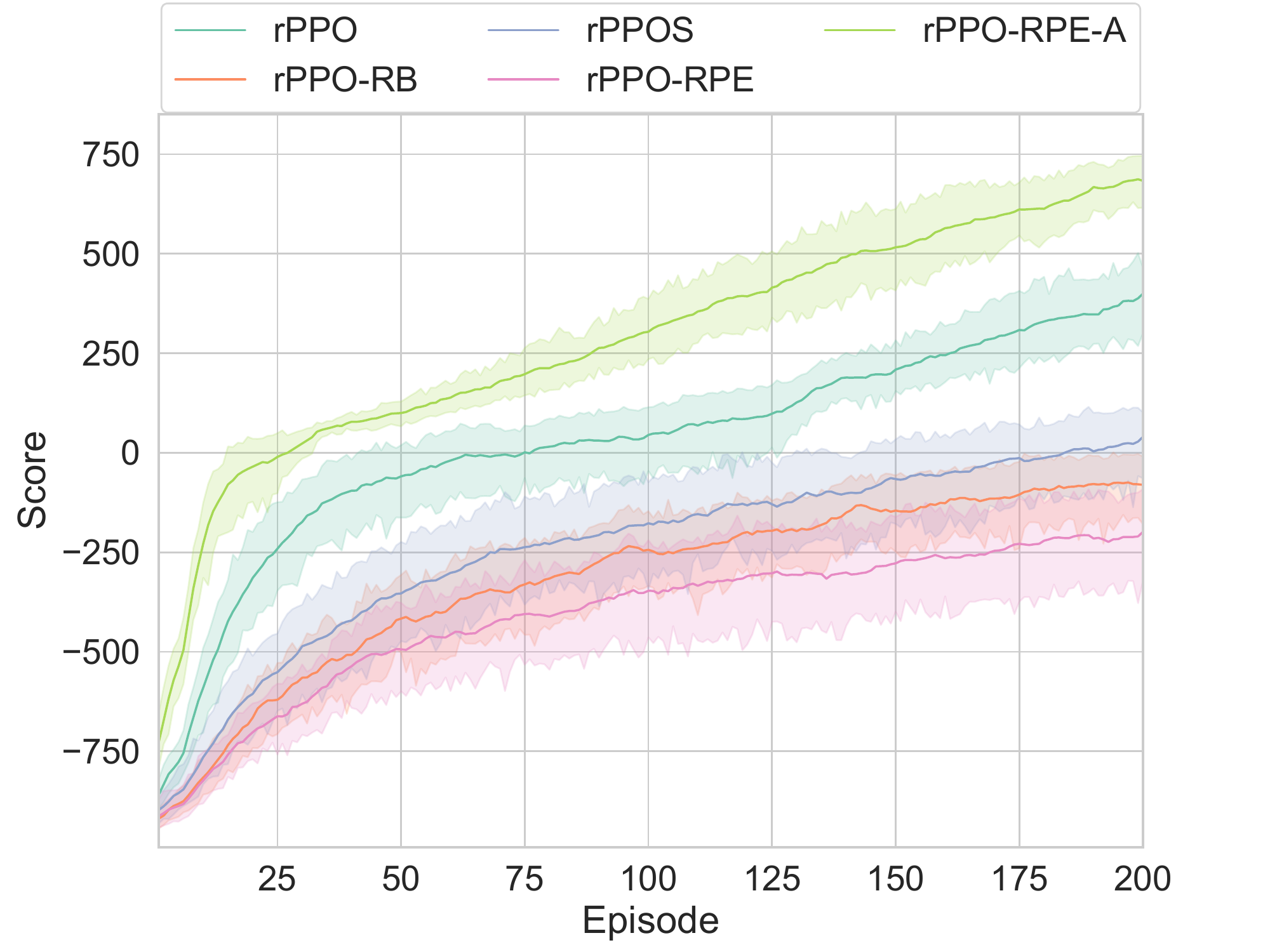}
    }
    \centering
    \subfigure[DoublePendulum w/ experience replay]{
        \includegraphics[keepaspectratio=true,width=0.315\linewidth]{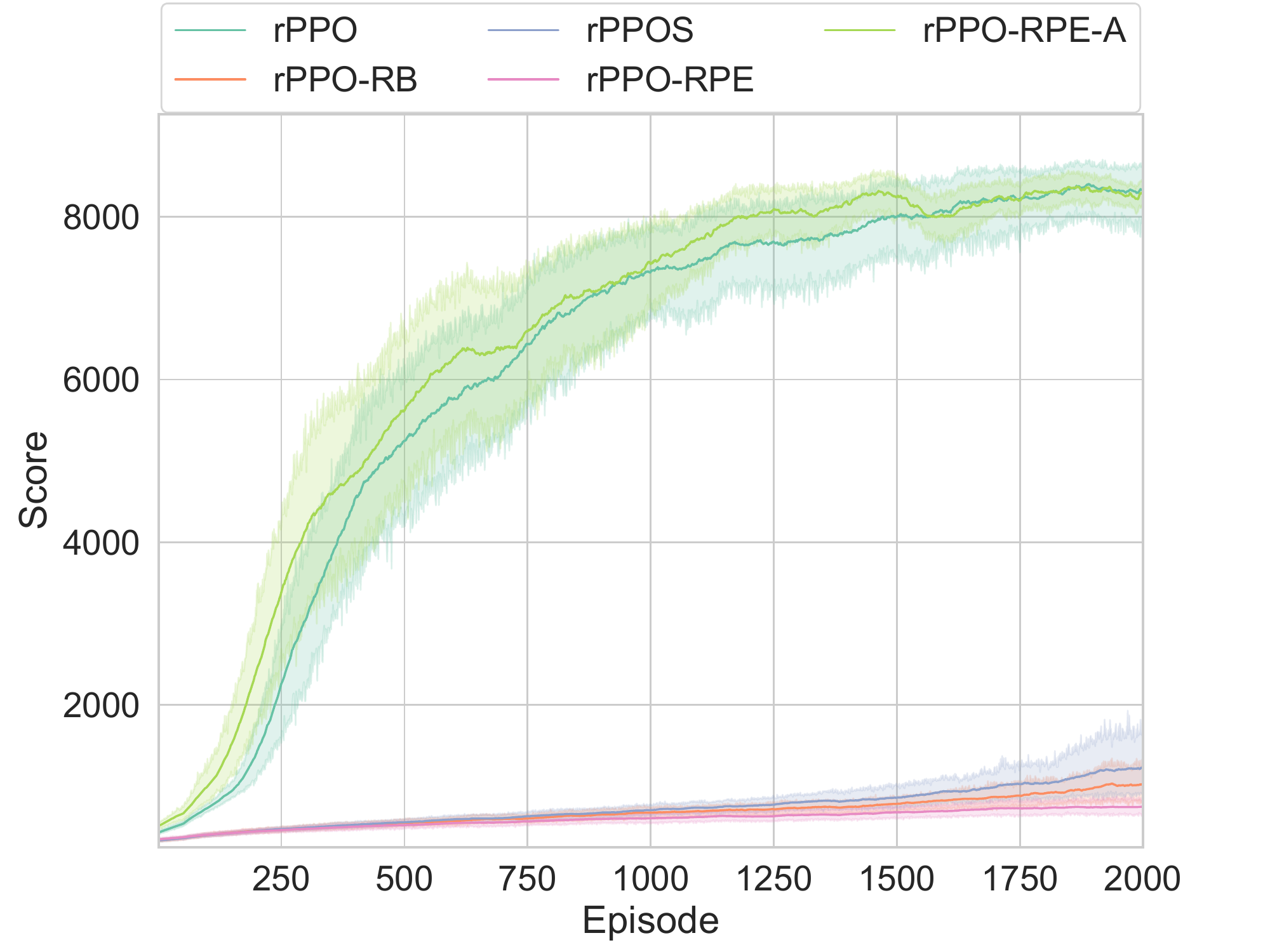}
    }
    \caption{Learning curves of three benchmark tasks:
        the sum of rewards at each episode are given as the score;
        the corresponding shaded areas show the 95~\% confidence intervals;
        in (a)--(c), the adaptive eligibility traces method~\citep{kobayashi2022adaptive} is employed, and due to the appropriate threshold, all the conditions succeeded in accomplishing InvertedPendulum and Swingup;
        however, due to the limitation of eligibility traces, DoublePendulum was accomplished only by PPO-RPE, and especially, the adaptive threshold enhanced the success rate of it;
        in (d)--(f), the experience replay method is employed, and due to too strict threshold, PPO-RB and PPO-RPE could not accomplish any task, although PPO could do so probably because of the lack of capability to softly constrain the latest policy to the baseline;
        by relaxation of the threshold, PPO-RPE-A achieved the best performance in all the tasks.
    }
    \label{fig:sims_learn}
\end{figure*}

\begin{figure*}[tb]
    \centering
    \subfigure[InvertedPendulum]{
        \includegraphics[keepaspectratio=true,width=0.315\linewidth]{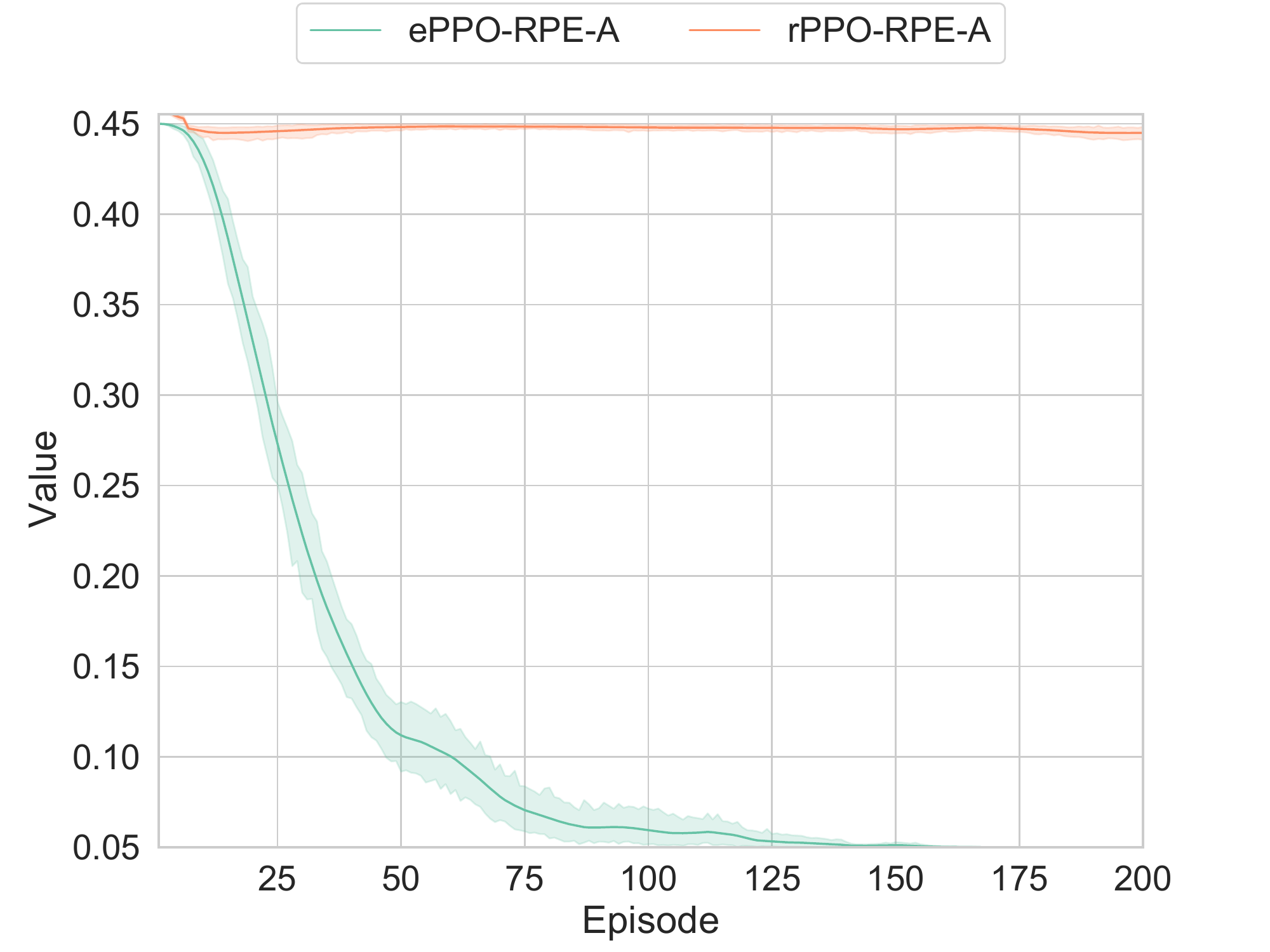}
    }
    \centering
    \subfigure[Swingup]{
        \includegraphics[keepaspectratio=true,width=0.315\linewidth]{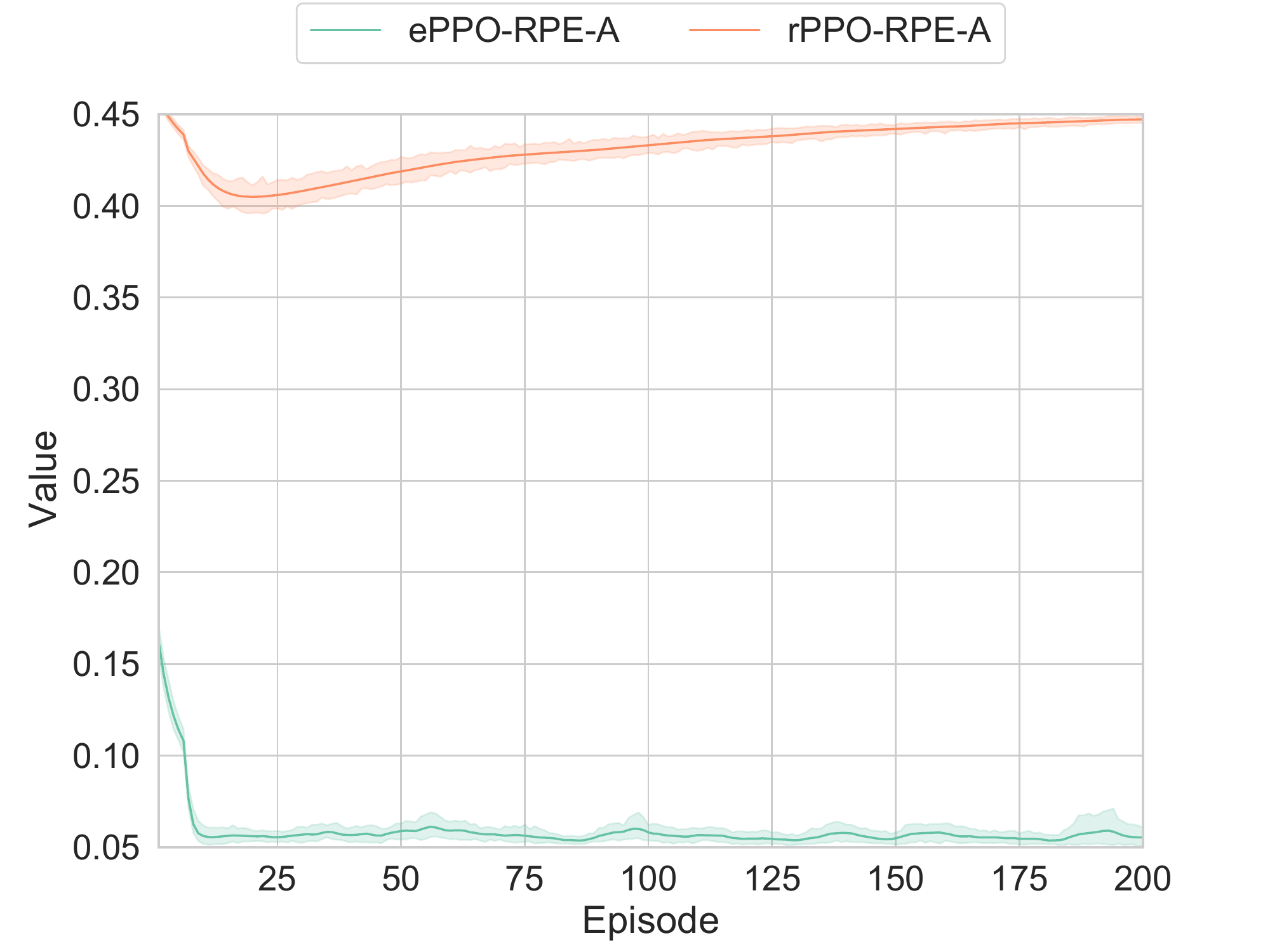}
    }
    \centering
    \subfigure[DoublePendulum]{
        \includegraphics[keepaspectratio=true,width=0.315\linewidth]{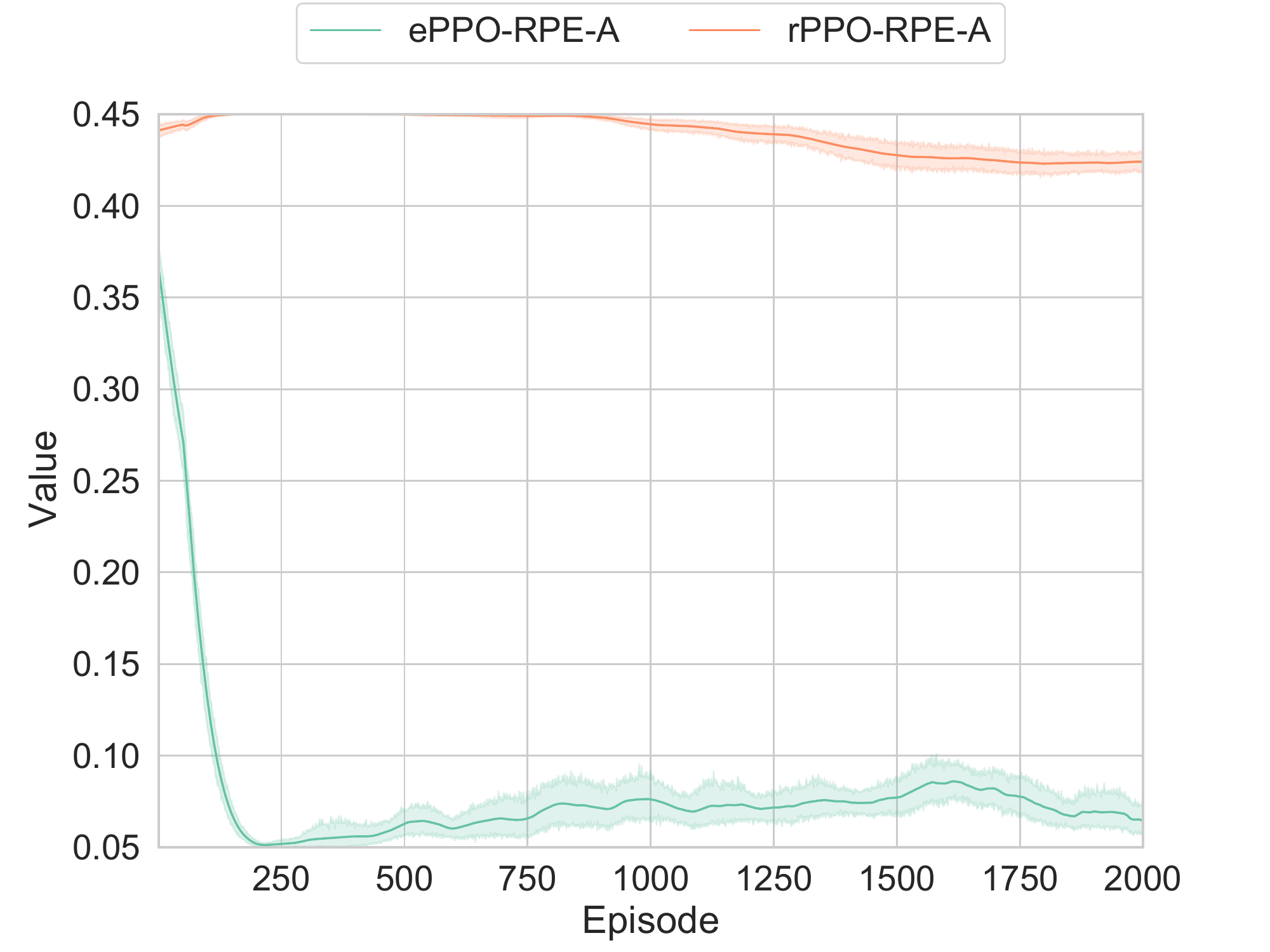}
    }
    \caption{Adaptive thresholds with two learning algorithms:
        the corresponding shaded areas show the 95~\% confidence intervals;
        with the eligibility traces, the value of $\epsilon$ was smoothly decreased to the near minimum because the latest and baseline policies ($\pi$ and $b$) are not far apart enough;
        with the experience replay, the value of $\epsilon$ was almost kept in the initial one (i.e. the maximum) since $b$ in the buffer is often old and $\pi$ tends to be far from it;
        however, in both cases, most of $\epsilon$ did not converge to a constant, and fluctuated adaptively according to the learning progress.
    }
    \label{fig:sims_threshold}
\end{figure*}

\subsection{Results for simple tasks}

To verify the value of the adaptive threshold, the first three tasks are conducted with the following conditions.
\begin{enumerate}
    \item \{e,r\}PPO~\citep{schulman2017proximal}: $\eta = 0$ as original PPO with no rollback
    \item \{e,r\}PPO-RB~\citep{wang2020truly}: $\eta = 0.3$ as the recommended value
    \item \{e,r\}PPOS~\citep{zhu2021functional}: $\eta = 0.3$ as the recommended value
    \item \{e,r\}PPO-RPE~\citep{kobayashi2021proximal}: the fixed threshold
    \item \{e,r\}PPO-RPE-A (proposal): the adaptive threshold
\end{enumerate}
where \{e,r\} denote the use of the adaptive eligibility traces or the experience replay.
For the fixed threshold, $\epsilon = 0.1$ is set as one of the recommended value and the same value as the earlier work~\citep{kobayashi2021proximal}, which is with the adaptive eligibility traces.
For the adaptive threshold, all the recommended values in Alg~\ref{alg:threshold} are utilized.
Note that the other PPO variants introduced in the introduction~\citep{hamalainen2020ppo,imagawa2019optimistic,libardi2021guided} were omitted because they are not for improvements of regularization and are difficult to compare fairly with the proposed method.

First of all, all the learning curves are illustrated in Fig.~\ref{fig:sims_learn}.
With the adaptive eligibility traces, InvertedPendulum and Swingup could be completely accomplished by all the conditions since the threshold $\epsilon$ was already tuned in the earlier work~\citep{kobayashi2021proximal}.
However, only PPO-RPE(-A) could accomplish DoublePendulum sometimes, and in particular, only the adaptive threshold enabled to reach the score around 5000, namely, it could succeeded in accomplishing DoublePendulum over 50~\%.
This is because online learning using the eligibility traces cannot maintain sufficient sample efficiency in failure-prone task such as DoublePendulum, but PPO-RPE limits the exploration range of the policy by regularizing the policy well, making it easier to find the optimal solution.

With the experience replay, PPO-RB, PPOS, and PPO-RPE failed to acquire all the tasks.
Since their performances generally decreased with stronger regularization methods, these results are probably due to too strong regularization of $\pi$ to $b$ for the replayed data.
As pointed out in~\citep{wang2020truly,zhu2021functional}, PPO has no capability to softly constrain $\pi$ to $b$ even with the too small threshold, and therefore, it yielded the near-optimal policy in the tasks except Swingup.
In contrast to them, only PPO-RPE-A stably learned all the tasks.
This is probably due to the relaxation of regularization (larger $\epsilon$ than 0.1).

To confirm the adaptability of the proposed method, Fig.~\ref{fig:sims_threshold} additionally shows the average thresholds in \{e,r\}PPO-RPE-A.
Basically, we can see that the adaptive eligibility traces and the experience replay required the small and large thresholds (i.e. around $\kappa \underline{\Delta}$ and $\kappa \overline{\Delta}$), respectively.
Note that clamping by $\underline{\Delta}$ and $\overline{\Delta}$ worked in a limited number of scenes, such as second half of learning in InvertedPendulum.
In addition, most of them did not converge to completely constant values, indicating its adaptive behavior according to the learning progress.
The benefits of such an adaptive behavior need to be investigated.

\subsection{Results for complex tasks}

\begin{figure}[tb]
    \centering
    \includegraphics[keepaspectratio=true,width=0.95\linewidth]{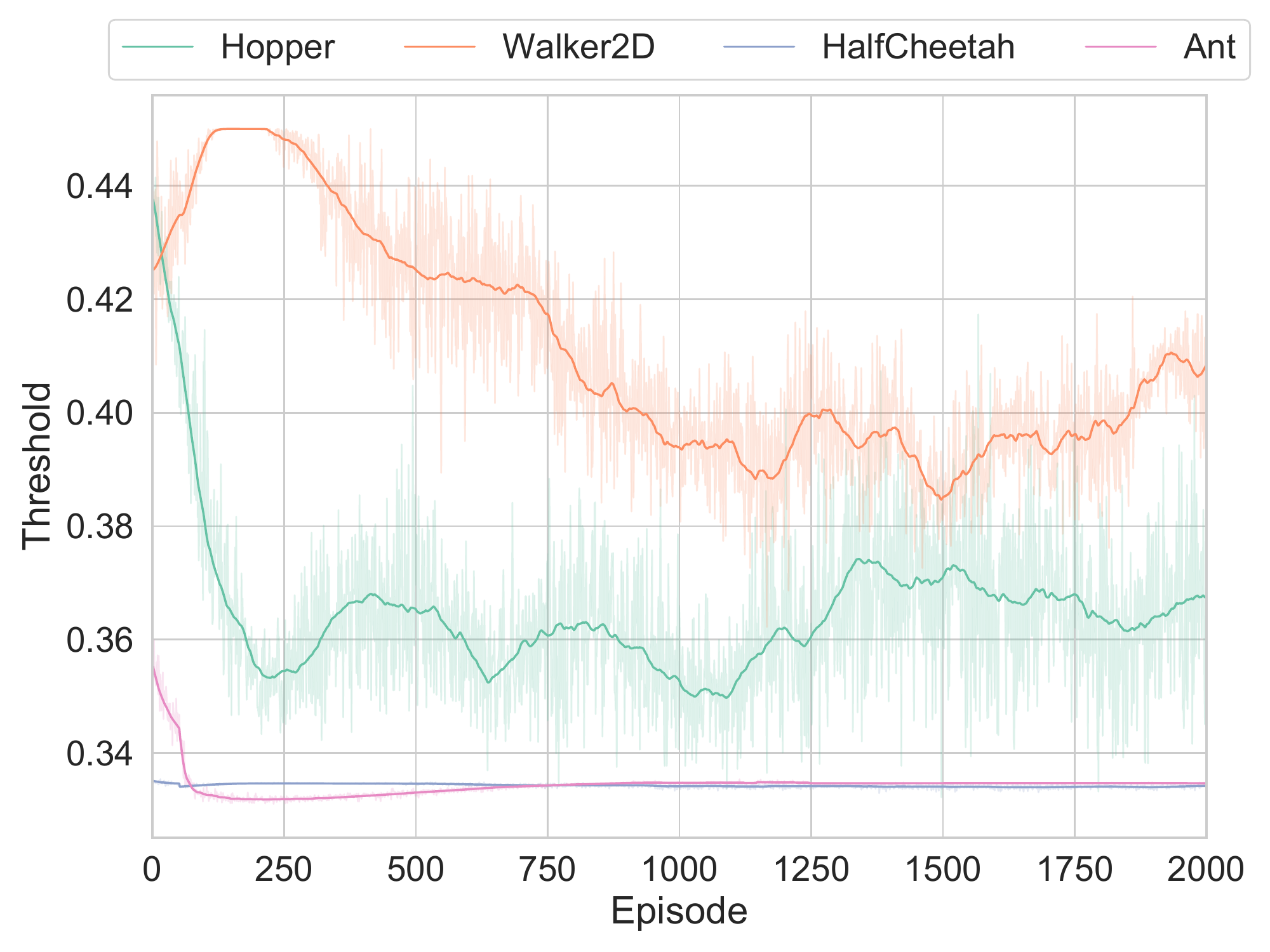}
    \caption{Trajectories of adaptive thresholds in the respective tasks:
        in stable HalfCheetah and Ant, their thresholds converged on almost the same values (i.e. $\sim 0.33$);
        in contrast, Hopper and Walker2D had oscillatory threshold due to their instability, which reduced the number of iterations for the policy updates.
    }
    \label{fig:simc_threshold}
\end{figure}

\begin{figure*}[tb]
    \centering
    \subfigure[Hopper]{
        \includegraphics[keepaspectratio=true,width=0.23\linewidth]{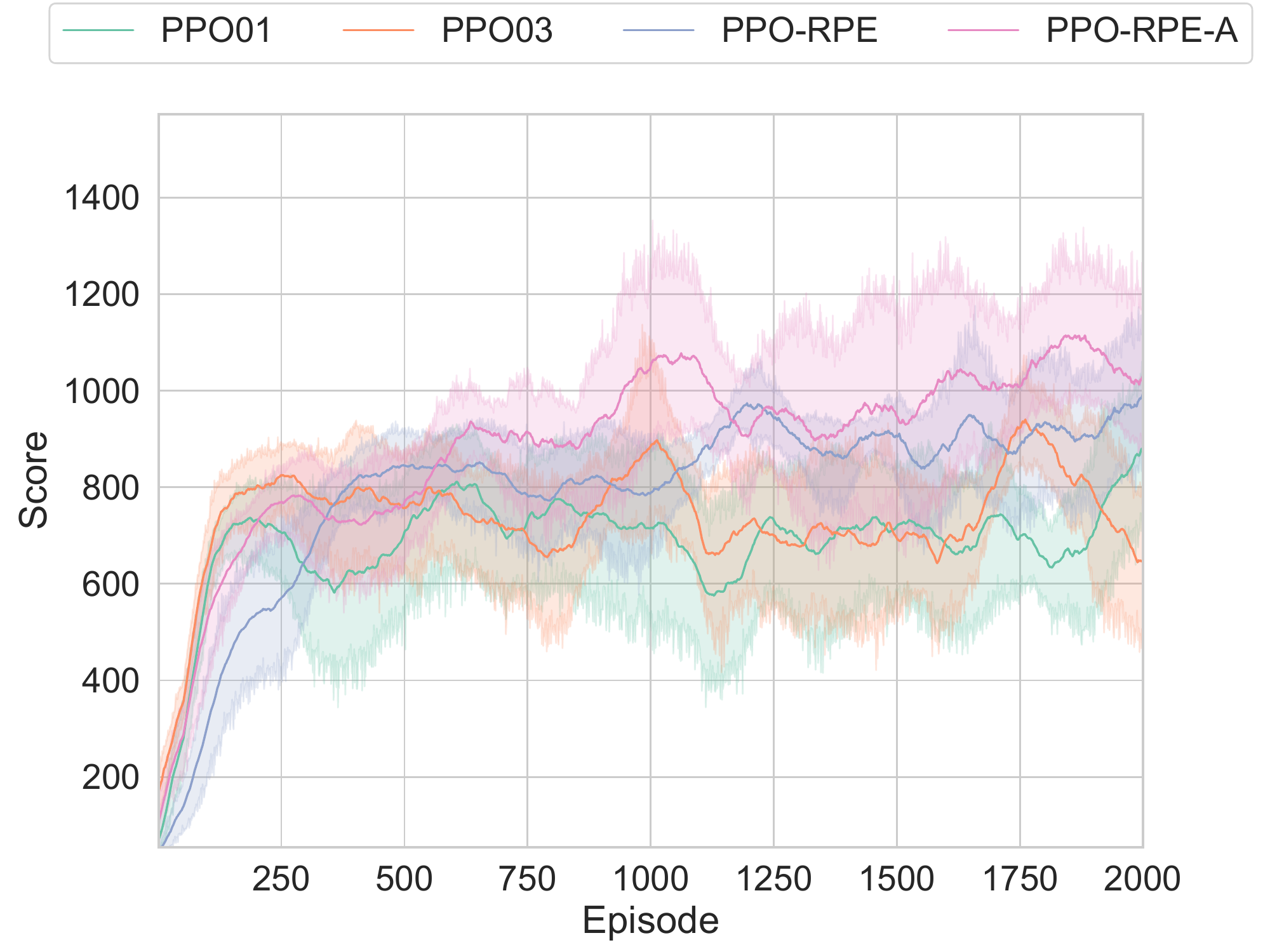}
    }
    \centering
    \subfigure[Walker2D (w/ stronger power)]{
        \includegraphics[keepaspectratio=true,width=0.23\linewidth]{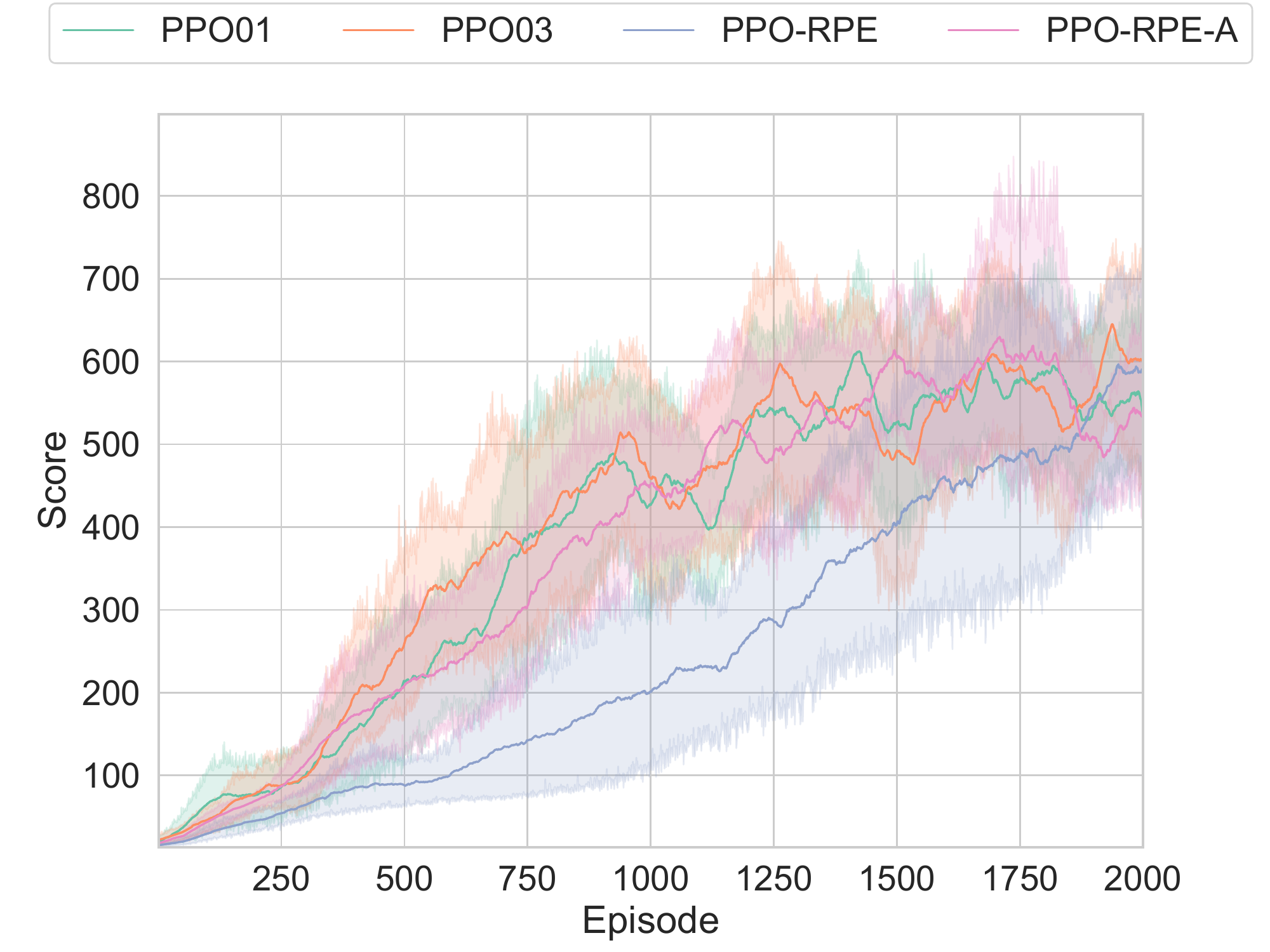}
    }
    \centering
    \subfigure[HalfCheetah]{
        \includegraphics[keepaspectratio=true,width=0.23\linewidth]{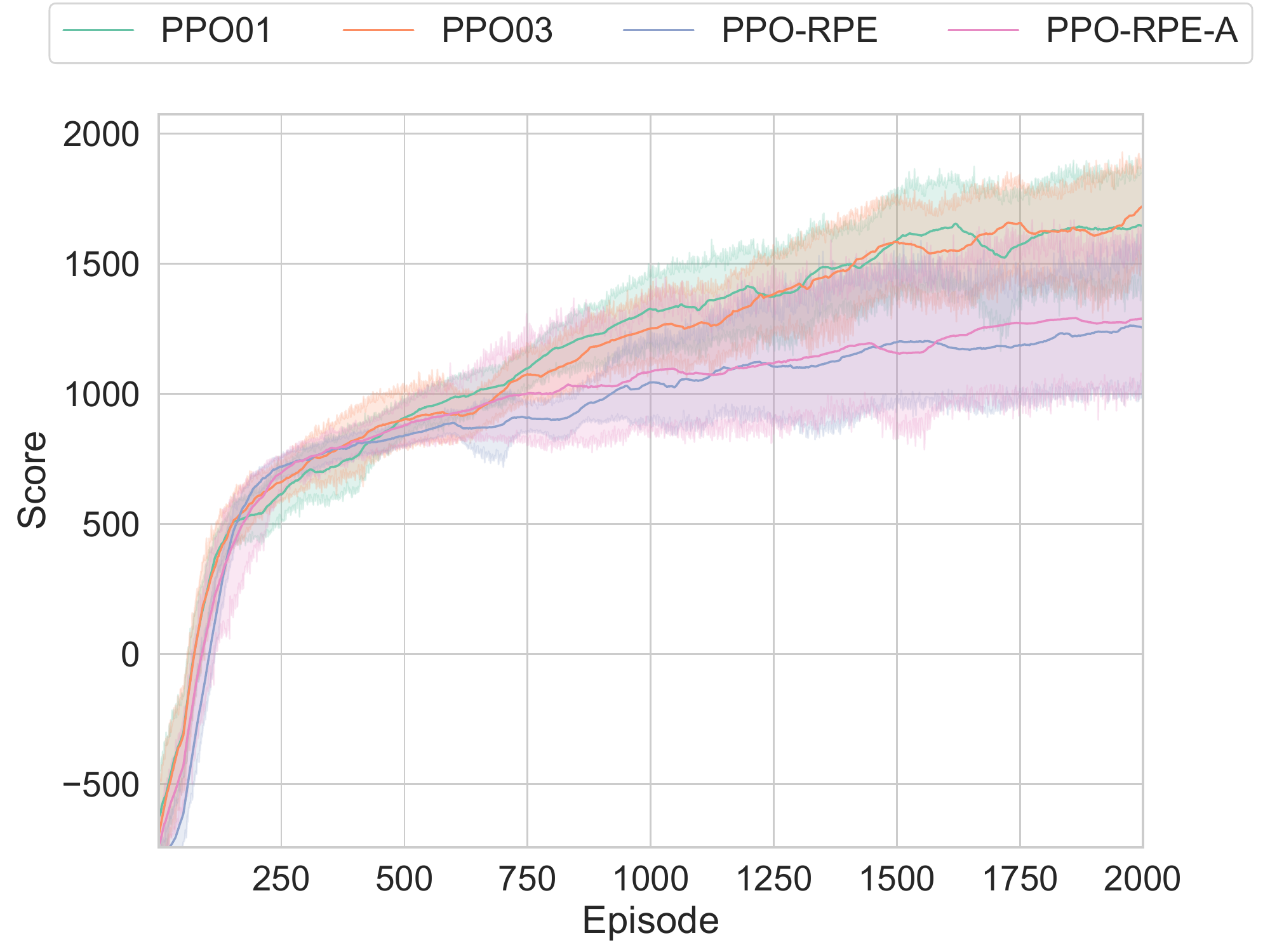}
    }
    \centering
    \subfigure[Ant]{
        \includegraphics[keepaspectratio=true,width=0.23\linewidth]{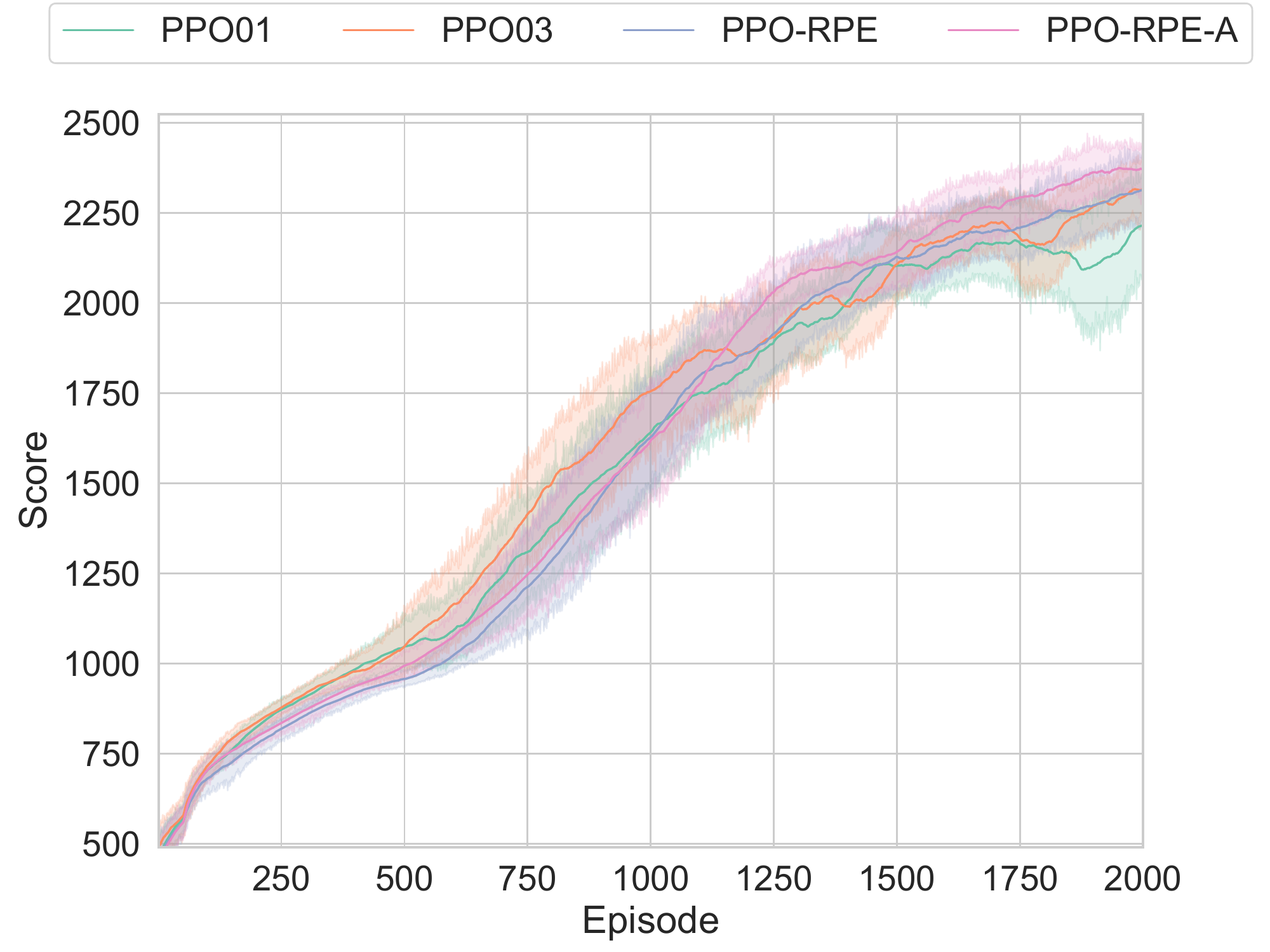}
    }
    \caption{Learning curves of four locomotion tasks:
        although (a) Hopper and (b) Walker2D tasks are with instability, which caused the larger confidence intervals, the proposed method, PPO-RPE-A, obtained the higher/same results than/as the others;
        in (c) HalfCheetah, only PPO succeeded in acquiring locomotion stably, and PPO-RPE failed probably due to the lack of the capability to explore a global solution;
        in contrast, (d) Ant probably requires only local exploration to find a global solution, and therefore, PPO-RPE-A, which can update the policy more carefully, eventually outperformed the others.
    }
    \label{fig:simc_learn}
\end{figure*}

\begin{figure}[tb]
    \centering
    \includegraphics[keepaspectratio=true,width=0.95\linewidth]{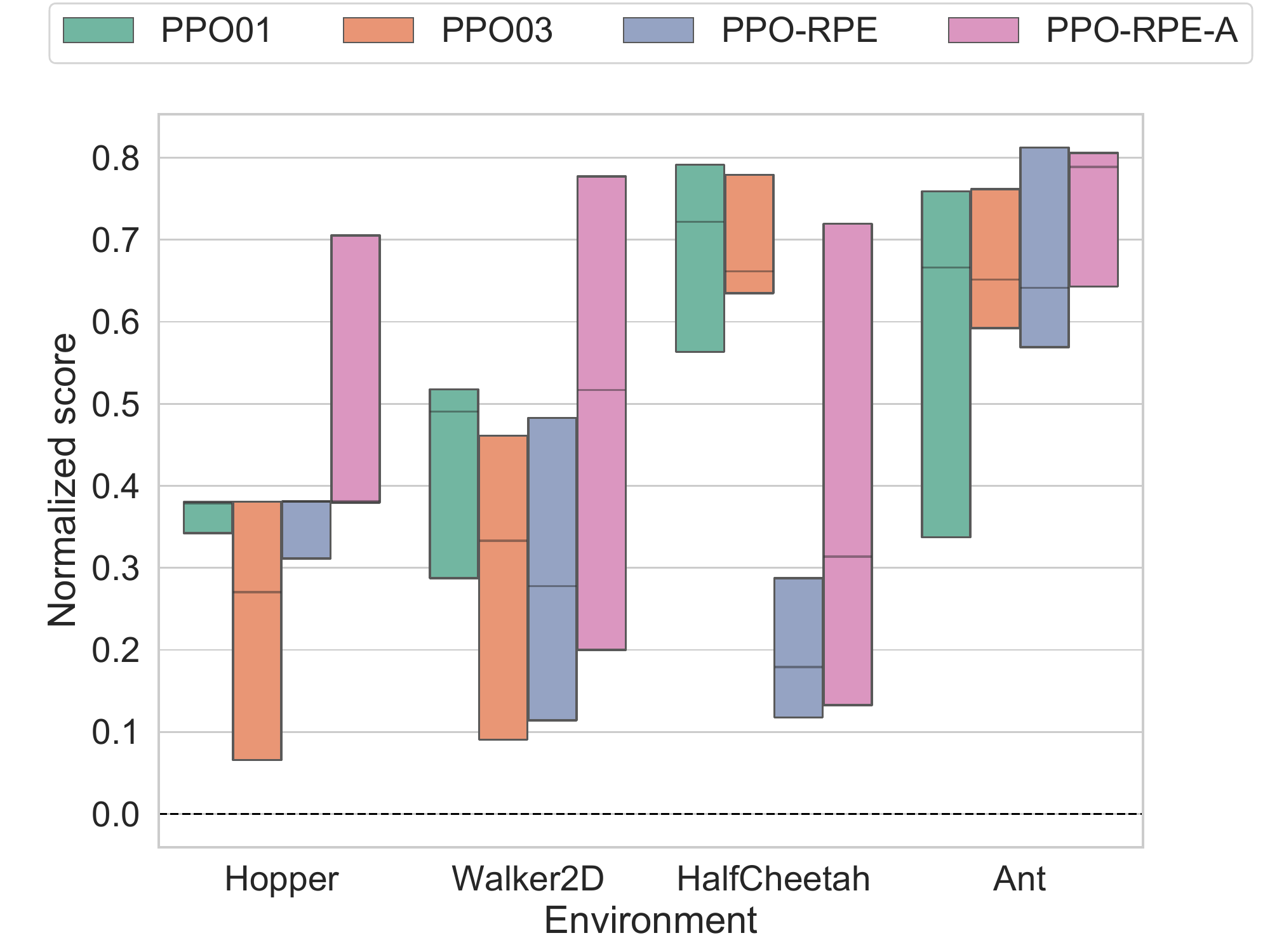}
    \caption{Test scores after training:
        in the unstable Hopper and Walker2D tasks, only the proposed method, PPO-RPE-A, succeeded in acquiring locomotion multiple times in 10 trials;
        PPO is suitable for learning HalfCheetah, which is stable and requires the high exploration capability, although the proposed method could adjust its regularization strength, yielding the multiple successes;
        in Ant task, PPO-RPE-A outperformed the others probably because it is stable and local exploration is enough to find a global solution.
    }
    \label{fig:simc_summary}
\end{figure}

\begin{figure*}[tb]
    \centering
    \subfigure[Hopper]{
        \includegraphics[keepaspectratio=true,width=0.23\linewidth]{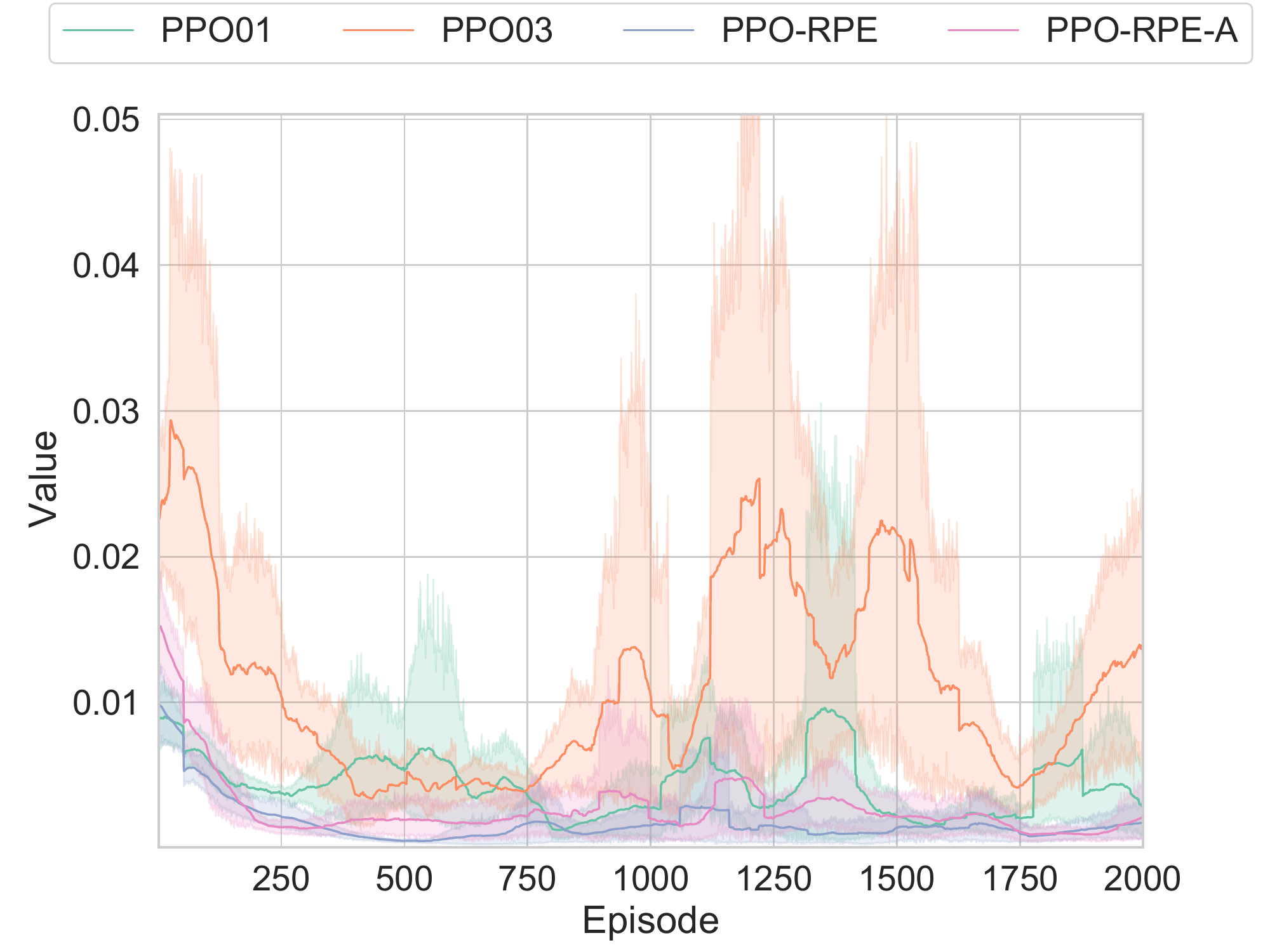}
    }
    \centering
    \subfigure[Walker2D (w/ stronger power)]{
        \includegraphics[keepaspectratio=true,width=0.23\linewidth]{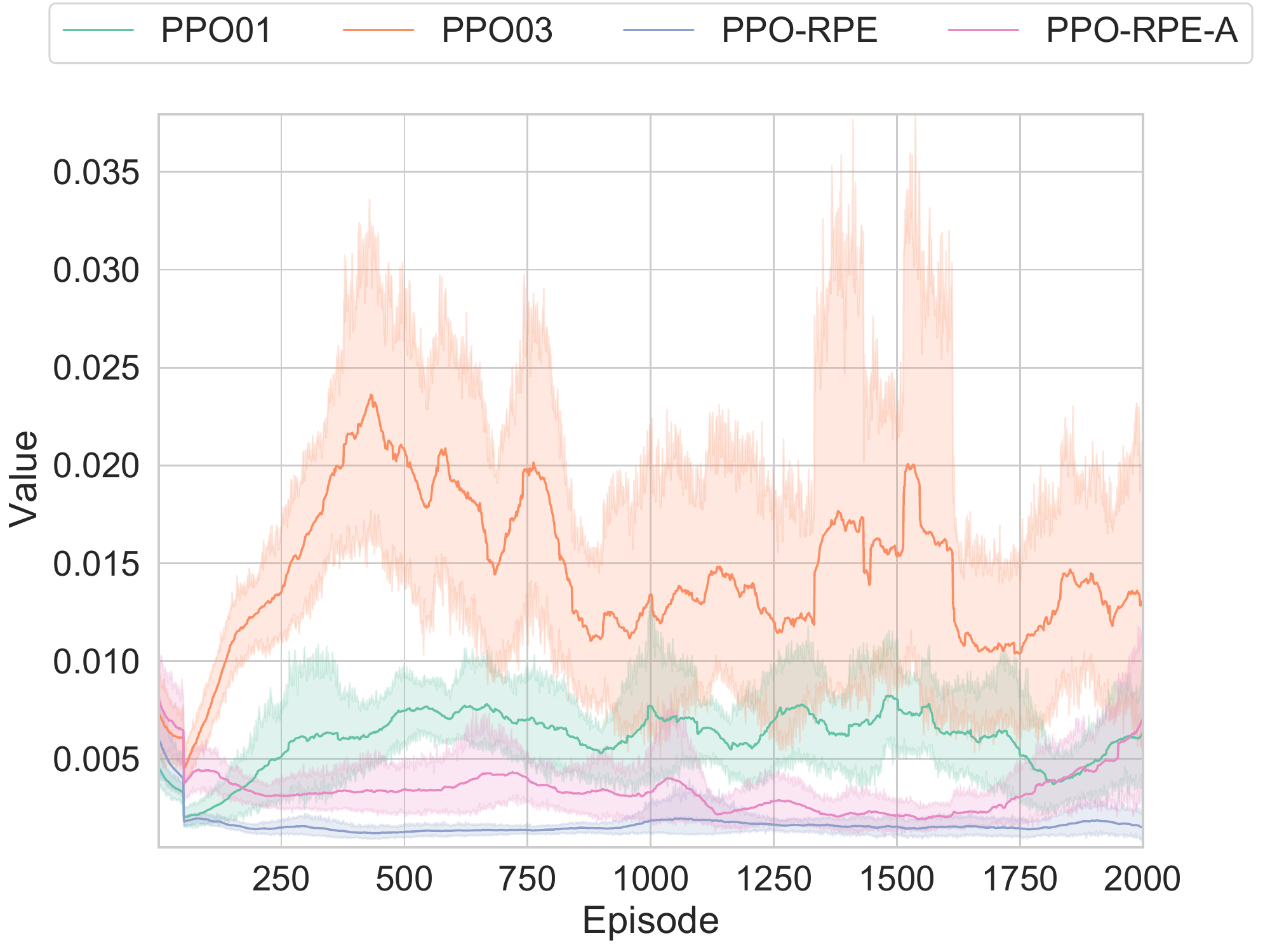}
    }
    \centering
    \subfigure[HalfCheetah]{
        \includegraphics[keepaspectratio=true,width=0.23\linewidth]{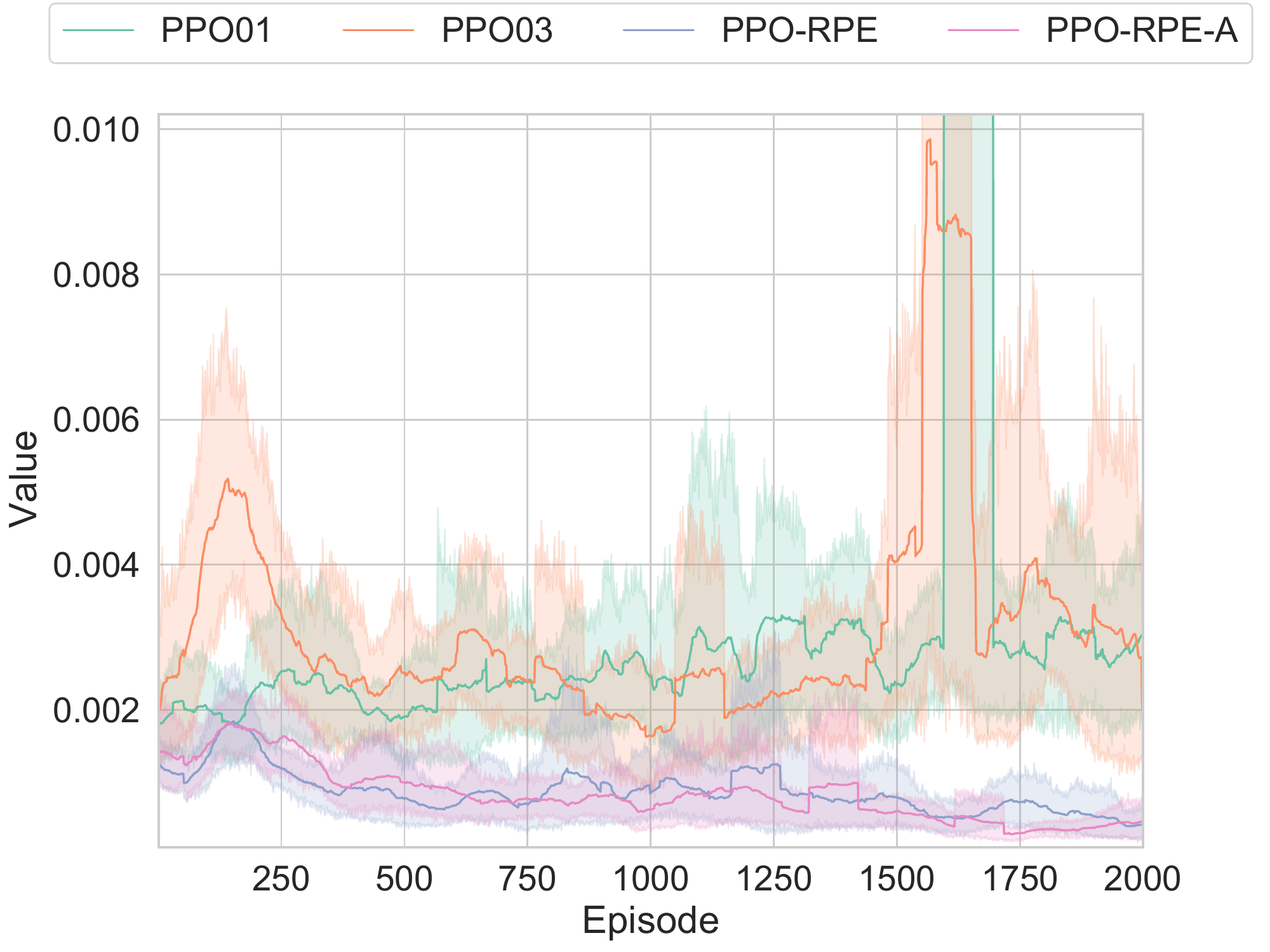}
    }
    \centering
    \subfigure[Ant]{
        \includegraphics[keepaspectratio=true,width=0.23\linewidth]{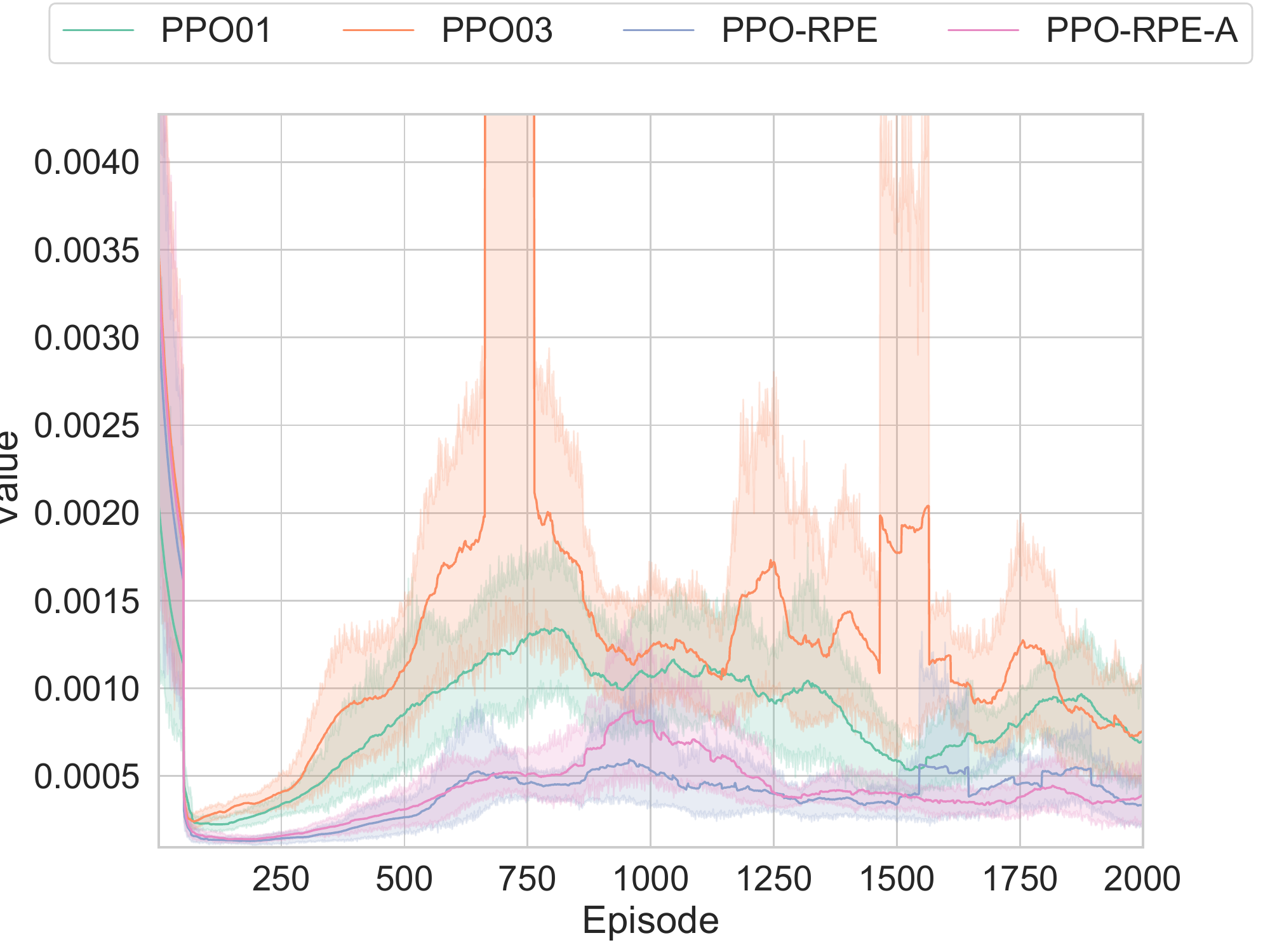}
    }
    \caption{Trajectories of numerically-approximated Pearson divergence between the latest and baseline policies:
        PPO obviously failed to softly constrain the latest policy to the baseline;
        both PPO-RPE with and without the adaptability of threshold served the appropriate regularization of the policy updates.
    }
    \label{fig:simc_diff}
\end{figure*}

For further investigation of the performance of the proposed method, the following experiment is also conducted on the remaining complex locomotion tasks under the following conditions.
\begin{enumerate}
    \item PPO01~\citep{schulman2017proximal}: $\epsilon = 0.1$
    \item PPO03~\citep{schulman2017proximal}: $\epsilon = 0.3$
    \item PPO-RPE~\citep{kobayashi2021proximal}: the fixed threshold $\epsilon = 0.3$
    \item PPO-RPE-A (proposal): the adaptive threshold
\end{enumerate}
where all the conditions are with both of the adaptive eligibility traces and the experience replay.
Due to the bad performance of PPO-RB in the previous results, it is omitted here.
To evaluate the value of the adaptive behavior in the learning progress, as indicated in the previous results, PPO-RPE (and PPO) with $\epsilon \simeq 0.3$, which is one of the recommended values closest to the approximate mean of the adaptive threshold (see Fig.~\ref{fig:simc_threshold}), are also compared.

All the learning curves are illustrated in Fig.~\ref{fig:simc_learn}.
In addition, after learning the policy for each task with each condition, the agent performed the task using the learned policy 100 times to compute the median of the scores.
This test results are summarized in Fig.~\ref{fig:simc_summary}.

The wide confidence intervals of the learning curves and the test scores suggest that Hopper and Walker2D are very unstable tasks, where the agent often encounters falling over and resetting the episode.
In such an environment, the policy should be updated carefully, and in fact, only the proposed method (i.e PPO-RPE-A) accomplished the task multiple times in 10 trials.
Note that PPO-RPE without the adaptive threshold could not obtain the same results because the appropriate regularization strength is not served.

On the other hand, in HalfCheetah, PPO with both thresholds outperformed the proposed method.
This is probably because this task is relatively stable, easily falling into a local solution (i.e. an upright state); as a result, sufficient exploration is required to get out of such a local solution.
As pointed out in~\citep{wang2020truly,zhu2021functional}, PPO does not softly constrain the latest policy to the baseline, and therefore, it may hold the exploration capability enough.
Nevertheless, it is clear that given the regularization strength adaptively, the task accomplishment rate by PPO-RPE-A is better than that by PPO-RPE.
Although Ant also seems to be a stable task, it only requires local exploration to improve the performance, hence, the proposed method obtained the maximum performance.

Finally, to reveal the gap between the latest and baseline policies, numerically-approximated Pearson divergence between the latest and baseline policies is depicted in Fig.~\ref{fig:simc_diff}.
Clearly, PPO fails to softly constrain the latest policy to the baseline, and is inadequate as a policy regularization method, although it does not impair the exploration capability.
In contrast, PPO-RPE succeeded in softly constraining the policy to the same degree, with or without a threshold.
This suggests that the adaptive threshold, which is adjusted at key points, contributes to the performance improvement, even though statistically giving the same level of regularization.

\section{Conclusion and future work}

This paper proposed PPO-RPE, a variant of PPO integrated with RPE divergence regularization, to clearly constrain the policy to its baseline in a symmetric manner.
The threshold-based gain was derived for PPO-RPE, as the standard PPO does, and makes this regularization tuning task-invariant.
In addition to the earlier work~\citep{kobayashi2021proximal}, this paper further contributes to design the adaptive threshold based on the symmetric property in the relative density ratio for RPE divergence.
The conventional threshold can be interpreted that it is set based on the theoretical error scale of the relative density ratio from its center, but the actual error scale is algorithm-dependent.
According to this fact, the adaptive threshold was designed to consider the algorithm-dependent error scale, which is estimated from experience heuristically.
This design can cancel the algorithm dependency of the regularization tuning.
In the numerical simulations with two types of learning algorithms, only the proposed method could accomplish all the tasks using both the learning algorithms, although the standard PPO was also robust to the change of learning algorithm due to the lack of the capability to softly constrain the latest policy to the baseline.
In the additional simulations for the four complex locomotion tasks, only the proposed method could achieve high scores even in  the unstable tasks (i.e. Hopper and Walker2D), although the firm regularization to the policy updates resulted in lower performance than PPO on the task requiring sufficient exploration (i.e. HalfCheetah).

The simulation results suggested that the regularization of the policy updates does not necessarily lead to improved performance of RL.
By appropriately adjusting $\kappa$ (e.g. decaying like~\citep{farsang2021decaying}), we can expect to get the better balance between the strength of the regularization and the exploration performance.
Alternatively, the existence of a flat region like PPO and PPOS may be useful in order not to deteriorate the exploration performance and/or not to constrain the latest policy to the too old baseline.
In the future, therefore, the proposed method will be improved in terms of the exploration performance while holding the regularization capability.
In addition, the proposed method can be integrated with the other PPO variants~\citep{hamalainen2020ppo,imagawa2019optimistic,libardi2021guided}.
Hence, by investigating the integration with them, we can expect to obtain better learning performance in the near future.

\appendix
\subsection{Regularization term in PPO and PPO-RB}
\label{app:omega_ppo}

With eqs.~\eqref{eq:loss_reg_ds},~\eqref{eq:ratio_ppo}, and~\eqref{eq:loss_ppo}, the regularization term for PPO, which has not been explicitly defined in the previous work~\citep{schulman2017proximal,wang2020truly} yet, is derived.
The changes in $\rho^\mathrm{PPO}$ compared to the original $\rho$ should be transferred into the surrogate advantage function $A^\mathrm{PPO}$, which is the sum of the original $A$ and the negative regularization term for PPO $- \Omega^\mathrm{PPO}$.
When no clipping or rollback is done, $\Omega^\mathrm{PPO}$ is clearly derived as zero.
In contrast, when clipping or rollback, the following relationship is derived.
\begin{align}
    \rho^\mathrm{PPO} A &= \rho A \{ - \eta + \rho^{-1} (1 + \eta) (1 + \sigma \epsilon) \}
    \nonumber \\
    &= \rho A \{ 1 - (1 + \eta) + \rho^{-1} (1 + \eta) (1 + \sigma \epsilon) \}
    \nonumber \\
    &= \rho A \{ 1 - (1 + \eta) (1 - \rho^{-1} (1 + \sigma \epsilon)) \}
    \nonumber \\
    &= \rho \{A - A (1 + \eta) (1 - \rho^{-1} (1 + \sigma \epsilon)) \}
    \nonumber \\
    &= \rho A^\mathrm{PPO} = \rho (A - \Omega^\mathrm{PPO})
\end{align}
Therefore, the regularization term in PPO, $\Omega^\mathrm{PPO}$, is given as follows:
\begin{align}
    \Omega^\mathrm{PPO} &= \begin{cases}
        A (1 + \eta) (1 - \rho^{-1} (1 + \sigma \epsilon)) & \sigma (\rho - 1) \geq \epsilon
        \\
        0 & \mathrm{otherwise}
    \end{cases}
\end{align}
As can be seen in this regularization term, PPO regularizes the policy
adaptively according to the advantage function $A$.
Although this seems to be a kind of hinge loss function, which is zero if and only if $\rho = 1 + \sigma \epsilon$ or no clipping and rollback, its mathematical meaning is not intuitive, even by converting it to its expectation $\mathbb{E}_b[\rho \Omega^\mathrm{PPO}]$ as the regularization target.

\subsection{Derivation of eq.~\eqref{eq:grad_rpe}}
\label{app:grad_rpe}

With $\rho_\beta = \rho / (1 - \beta + \beta \rho)$, the partial differentiation of $\partial \rho_\beta / \partial \rho$ is given as follows:
\begin{align}
    \cfrac{\partial \rho_\beta}{\partial \rho} &= \cfrac{1 - \beta + \beta \rho - \beta \rho}{(1 - \beta + \beta \rho)^2}
    = \cfrac{1 - \beta}{(1 - \beta + \beta \rho)^2}
\end{align}
Using this, eq.~\eqref{eq:grad_rpe} can be derived in the following steps.
\begin{align}
    \tilde{A}^\mathrm{RPE} &= A \!-\! C \beta (\rho_\beta - 1)^2 \!-\! 2 C (1 - \beta + \beta \rho) (\rho_\beta - 1) \cfrac{\partial \rho_\beta}{\partial \rho}
    \nonumber \\
    &= A \!-\! C \beta (\rho_\beta - 1)^2 \!-\! C (\rho_\beta - 1) \cfrac{2(1 - \beta + \beta \rho) (1 - \beta)}{(1 - \beta + \beta \rho)^2}
    \nonumber \\
    &= A - C \beta (\rho_\beta - 1)^2 - C (\rho_\beta - 1) \cfrac{2(1 - \beta)}{1 - \beta + \beta \rho}
    \nonumber \\
    &= A - C (\rho_\beta - 1) \left \{ \beta (\rho_\beta  - 1) + \cfrac{2(1 - \beta)}{1 - \beta + \beta \rho} \right \}
    \nonumber \\
    &= A \left [ 1 - \cfrac{C}{A}(\rho_\beta - 1)\left \{ \beta (\rho_\beta  - 1) + \cfrac{2 (1 - \beta)}{1 - \beta + \beta \rho} \right \} \right ]
    \nonumber \\
    &= A \left [ 1 - \cfrac{C}{A}(\rho_\beta - 1)\left \{ \beta (\rho_\beta  - 1) + 2 (1 - \beta) \cfrac{\rho_\beta}{\rho} \right \} \right ]
\end{align}
where $1 / (1 - \beta + \beta \rho) = \rho_\beta / \rho$ is utilized for the last line of the equation transformation.

\section*{Acknowledgments}

This work was supported by JSPS KAKENHI, Grant-in-Aid for Scientific Research (B), Grant Number JP20H04265.

\bibliographystyle{elsarticle-harv}
\bibliography{pporpe}

\end{document}